%% file: arxiv-main.tex
\documentclass[sigconf]{acmart}

\copyrightyear{2018}
\acmYear{2018} 
\acmConference[WWW 2018]{The 2018 Web Conference}{April 23--27, 2018}{Lyon, France}
\acmPrice{}
\acmDOI{https://doi.org/10.1145/3178876.3186071}
\acmISBN{978-1-4503-5639-8}

\acmArticle{4}

\usepackage{balance}
\usepackage{amsmath}
\usepackage{amsthm}
\usepackage{amsfonts}
\usepackage{amssymb}
\usepackage{adjustbox}
\usepackage{etex}
\usepackage{subfigure}
\usepackage{graphicx}
\usepackage{bm}
\usepackage{url}
\usepackage{stmaryrd}
\usepackage{balance}
\usepackage{epstopdf}
\usepackage{multirow}
\usepackage{booktabs}
\graphicspath{{./figures/}} 
\usepackage{seqsplit}

 \usepackage{pgfplots}
\usepackage{pifont}
\usepackage{epsfig}
 \usepackage{pgffor}
 \usepackage{tikz}

\begin{document}
\title{Learning from Multi-View Multi-Way Data via Structural Factorization Machines}
\renewcommand{\shorttitle}{Structural Factorization Machines}

\author{Chun-Ta Lu}
\affiliation{%
       \institution{University of Illinois at Chicago}
}
\email{clu29@uic.edu}

\author{Lifang He}
\authornote{Corresponding author.}
\affiliation{%
       \institution{Cornell University}
}
\email{lifanghescut@gmail.com}

\author{Hao Ding}
\affiliation{%
       \institution{Purdue University}
}
\email{haoding.tourist@gmail.com}

\author{Bokai Cao}
\affiliation{%
    \institution{University of Illinois at Chicago}
}
\email{caobokai@uic.edu}

\author{Philip S. Yu}
\affiliation{%
       \institution{University of Illinois at Chicago}
}
\affiliation{%
       \institution{Tsinghua University}
}
\email{psyu@cs.uic.edu}

\renewcommand{\shortauthors}{C.-T. Lu et al.}

\begin{abstract}
Real-world relations among entities can often be observed and determined by different perspectives/views. 
For example, the decision made by a user on whether to adopt an item relies on multiple aspects such as the contextual information of the decision, the item's attributes, the user's profile and the reviews given by other users. Different views may exhibit multi-way interactions among entities and provide complementary information.
In this paper, we introduce a multi-tensor-based approach that can preserve the underlying structure of multi-view data in a generic predictive model. Specifically, we propose structural factorization machines (SFMs) that learn the common latent spaces shared by multi-view tensors and automatically adjust the importance of each view in the predictive model. Furthermore, the complexity of SFMs is linear in the number of parameters, which make SFMs suitable to large-scale problems. Extensive experiments on real-world datasets demonstrate that the proposed SFMs outperform several state-of-the-art methods in terms of prediction accuracy and computational cost.
\end{abstract}

\ccsdesc[500]{Computing methodologies~Machine learning}
\ccsdesc[500]{Computing methodologies~Supervised learning}
\ccsdesc[500]{Computing methodologies~Factorization methods}

\keywords{Tensor Factorization; Multi-Way Interaction; Multi-View Learning}

\maketitle

\input{01-intro}
\input{06-related}
\input{02-prelim}
\input{03-model}
\input{04-learning}
\input{05-exp}
\input{07-conclusion}

\section*{Acknowledgments}
This work is supported in part by NSF through grants IIS-1526499, and CNS-1626432, and NSFC 61672313, 61503253 and NSF of Guangdong Province (2017A030313339). We gratefully acknowledge the support of NVIDIA Corporation with the donation of the Titan X GPU used for this research.

\balance
\bibliographystyle{acm}
\bibliography{reference}

\end{document}

%% file: 01-intro.tex
\section{Introduction}


With the ability to access massive amounts of heterogeneous data from multiple sources, multi-view data have become prevalent in many real-world applications. For instance, in recommender systems, online review sites (like Amazon and Yelp) have access to contextual information of shopping histories of users, the reviews written by the users, the categorizations of the items, as well as the friends of the users. 
Each view may exhibit pairwise interactions (e.g., the friendships between users) or even higher-order interactions (e.g., a customer write a review for a product) among entities (such as customers, products, and reviews), and can be represented in a multi-way data structure, i.e., tensor. 
Since different views usually provide complementary information~\cite{cao2014tensor,cao2016multi,lu2017mfm}, how to effectively incorporate information from multiple structural views is critical to good prediction performance for various machine learning tasks. 

Typically, a predictive model is defined as a function of predictor variables (e.g., the customer id, the product id, and the categories of the product) to some target (e.g., the rating). The most common approach in predictive modeling for multi-view multi-way data is to describe samples with feature vectors that are flattened and concatenated from structural views, and apply a vector-based method, such as linear regression (LR) and support vector machines (SVMs), to learn the target function from observed samples. Recent works have shown that linear models fail for tasks with very sparse data~\cite{rendle2012factorization}. A variety of methods have been proposed to address the data sparsity issue by factorizing the monomials (or feature interactions) with kernels, such as the ANOVA kernels used in FMs~\cite{rendle2012factorization,blondel2016higher} and polynominal kernels used in polynominal networks~\cite{livni2014computational,blondelIFU16}. 
However, the disadvantages of this approach are that (1) the important structural information of each view will be discarded which may lead to the degraded prediction performance and (2) the feature vectors can grow very large which can make learning and prediction very slow or even infeasible, especially if each view involves relations of high cardinality. For example, including the relation ``friends of a user'' in the feature vector (represented by their IDs) can result in a very long feature vector. Further, it will repeatedly appear in many samples that involve the given user. 

Matrix/tensor factorization models have been a topic of interest in the areas of multi-way data analysis, e.g., community detection~\cite{he2016joint}, collaborative filtering~\cite{koren2008factorization,rendle2010pairwise}, knowledge graph completion~\cite{zhang2017connecting}, and neuroimage analysis~\cite{he2014dusk}. Assuming multi-view data have the same underlying low-rank structure (at least in one mode), coupled data analysis such as collective matrix factorization (CMF)~\cite{singh2008relational} and coupled matrix and tensor factorization (CMTF)~\cite{acar2011all} that jointly factorize multiple matrices (or tensors) has been applied to applications such as clustering and missing data recovery. However, they are only applicable to categorical variables. Moreover, since existing coupled factorization models are unsupervised, the importance of each structural view in modeling the target value cannot be automatically learned. 
Furthermore, when applying these models to data with rich meta information (e.g., friendships) but extremely sparse target values (e.g., ratings), it is very likely the learning process will be dominated by the meta information without manual tuning some hyperparameters, e.g., the weights of the fitting error of each matrix/tensor in the objective function~\cite{singh2008relational}, the weights of different types of latent factors in the predictive models~\cite{koren2010factor}, or the regularization hyperparamters of latent factor alignment~\cite{lu2016item}. 

In this paper, we propose a general and flexible framework for learning the predictive structure from the complex relationships within the multi-view multi-way data. 
Each view of an instance in this framework is represented by a tensor that describes the multi-way interactions of subsets of entities, and different views have some entities in common. 
Constructing the tensors for each instance may not be realistic for real-world applications in terms of space and computational complexity, and the model parameters can have exponential growth and tend to be overfitting. 
In order to preserve the structural information of multi-view data without physically constructing the tensors, we introduce structural factorization machines (SFMs) that can learn the consistent representations in the latent feature spaces shared in the multi-view tensors while automatically adjust the contribution of each view in the predictive model. 
Furthermore, we provide an efficient method to avoid redundant computing on repeating patterns stemming from the relational structure of the data, such that SFMs can make the same predictions but with largely speed up computation. 

The contributions of this paper are summarized as follows:
\begin{itemize} \vspace{-2mm}
    \item We introduce a novel multi-tensor framework for mining data from heterogeneous domains, which can explore the high order correlations underlying multi-view multi-way data in a generic predictive model.
    \item We develop structural factorization machines (SFMs) tailored for learning the common latent spaces shared in multi-view tensors and automatically adjusting the importance of each view in the predictive model. The complexity of SFMs is linear in the number of features, which makes SFMs suitable to large-scale problems. 
    \item Extensive experiments on eight real-world datasets are performed along with comparisons to existing state-of-the-art factorization models to demonstrate its advantages. 
\end{itemize}

The rest of this paper is organized as follows. In Section~\ref{sec:related}, we briefly review related work on factorization models and multi-view learning. We introduce the preliminary concepts and problem definition in Section~\ref{sec:prelim}. We then propose the framework for learning multi-view multi-way data, and develop the structural factorization machines (SFMs), and provide an efficient computing method  in Section~\ref{sec:model}. The experimental results and parameter analysis are reported in Section~\ref{sec:exp}. 
Section~\ref{sec:conclusion} concludes this paper.

%% file: 06-related.tex
\section{Related Work}\label{sec:related}

\textbf{Feature Interactions.} 
Rendle pioneered the concept of feature interactions in Factorization Machines (FM) \cite{rendle2012factorization}. 
Juan et al. presented Field-aware Factorization Machines (FFM) \cite{juan2016field} to allow each feature to interact differently with another feature depending on its field.
Novikov et al. proposed Exponential Machines (ExM) \cite{novikov2016exponential} where the weight tensor is represented in a factorized format called Tensor Train. 
Zhang et al. used FM to initialize the embedding layer in a deep model \cite{zhang2016deep}.
Qu et al. added a product layer on the top of the embedding layer to increase the model capacity \cite{qu2016product}.
Other extensions of FM to deep architectures include Neural Factorization Machines (NFM) \cite{he2017neural} and Attentional Factorization Machines (AFM) \cite{xiao2017attentional}.
In order to effectively model feature interactions, a variety of models has been developed in the industry as well. Microsoft studied feature interactions in deep models, including Deep Semantic Similarity Model (DSSM) \cite{huang2013learning}, Deep Crossing \cite{shan2016deep} and Deep Embedding Forest \cite{zhu2017deep}. They use features as raw as possible without manually crafted combinatorial features, and let deep neural networks take care of the rest. Alibaba proposed a Deep Interest Network (DIN) \cite{zhou2017deep} to learn user embeddings as a function of ad embeddings. Google used deep neural networks to learn from heterogeneous signals for YouTube recommendations \cite{covington2016deep}.
In addition, Wide \& Deep Models \cite{cheng2016wide} were developed for app recommender systems in Google Play where the wide component includes cross features that are good at memorization and the deep component includes embedding layers for generalization. Guo et al. proposed to use FM as the wide component in Wide \& Deep with shared embeddings in the deep component \cite{guo2017deepfm}. Wang et al. developed the Deep \& Cross Network (DCN) to learn explicit cross features of bounded degree \cite{wang2017deep}.

\textbf{Multi-View Learning.} 
Multi-view learning (MVL) is concerned with predicting unknown values by taking multiple views into account. The traditional MVL refers to using relational features to construct a set of disjoint views, and these uncorrelated views are then used to model a target function to approximate the target concept to be learned~\cite{guo2006mining}. There are currently a plethora of studies available for MVL. Interested readers are referred to~\cite{xu2013survey} for a comprehensive survey of these techniques and applications. The most related works to ours are~\cite{cao2016multi,cao2017deepmood,liang2017icdm} that introduced and explored the tensor product operator to integrate different views together in a  tensor. Lu et al. further studied the multi-view feature interactions in the context of multi-task learning \cite{lu2017mfm}. 
However, this approach will introduce unexpected noise from the irrelevant feature interactions that can even be exaggerated after combinations, thereby degrading performance as demonstrated in the experiments. Different from conventional MVL approaches, the proposed algorithm can learn the common latent spaces shared in multi-view tensors and automatically adjusting the importance of each view in the predictive model.  

%% file: 02-prelim.tex
\section{Preliminaries}\label{sec:prelim}
In this section, we begin with a brief introduction to some related concepts and notation in tensor algebra, and then proceed to formulate the problem we are concerned with multi-view learning.

\begin{figure*}[t]
\centering
      \includegraphics[width=1.8\columnwidth]{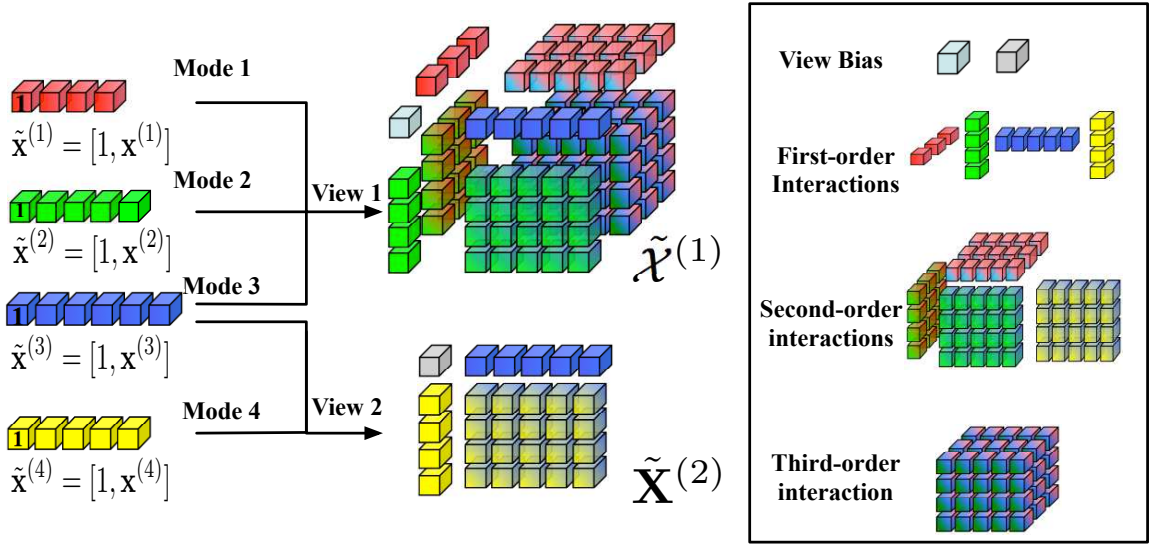}
  \caption{Example of multiple structural views, where  $\tilde{\mathcal{X}}^{(1)}= \tilde{\mathbf{x}}^{(1)}\circ \tilde{\mathbf{x}}^{(2)} \circ \tilde{\mathbf{x}}^{(3)}$ and $\tilde{\mathbf{X}}^{(2)}= \tilde{\mathbf{x}}^{(3)}\circ \tilde{\mathbf{x}}^{(4)}$. 
  }\label{fig:input}
\end{figure*}
\subsection{Tensor Basics and Notation}
Tensor is a mathematical representation of a multi-way array. The order of a tensor is the number of modes (or ways). A zero-order tensor is a scalar, a first-order tensor is a vector, a second-order tensor is a matrix and a tensor of order three or higher is called a higher-order tensor. 
An element of a vector $\mathbf{x}$, a matrix $\mathbf{X}$, or a tensor $\mathcal{X}$ is denoted by $x_i$, $x_{i,j}$, $x_{i,j,k}$, etc., depending on the number of modes. All vectors are column vectors unless otherwise specified. For an arbitrary matrix $\mathbf{X} \in \mathbb{R}^{I \times J}$, its $i$-th row and $j$-th column vector are denoted by $\mathbf{x}^{i}$ and $\mathbf{x}_{j}$, respectively. Given two matrices $\mathbf{X}, \mathbf{Y} \in \mathbb{R}^{I \times J}$,
$\mathbf{X} * \mathbf{Y}$ denotes the element-wise (Hadamard) product between $\mathbf{X}$ and $\mathbf{Y}$, defined as the matrix in $\mathbb{R}^{I \times J}$. 
An overview of the basic symbols used in this paper can be found in Table~\ref{tab:notation}.
\begin{definition}[Inner product] The inner product of two same-sized tensors $\mathcal{X}, \mathcal{Y} \in \mathbb{R}^{I_{1} \times I_{2} \times \cdots \times I_{M}}$ is defined as the sum of the products of their entries:
\begin{equation}\label{eq1}
\left\langle \mathcal {X}, \mathcal {Y}\right\rangle=\sum_{i_{1}=1}^{I_1}\sum_{i_{2}=1}^{I_2}\cdots\sum_{i_{M}=1}^{I_M} x_{i_1,i_2,\ldots,i_M} y_{i_1,i_2,\ldots,i_M}.
\end{equation}
\end{definition}


\input{table_notation}


\begin{definition}[Outer product] 
The outer product of two tensors $\mathcal{X} \in \mathbb{R}^{I_{1} \times I_{2} \times \cdots \times I_{N}}$ and $\mathcal{Y}\in \mathbb{R}^{I_{1}' \times I_{2}' \times \cdots \times I_{M}'}$ is a $(N+M)$th-order tensor denoted by $\mathcal {X} \circ \mathcal {Y}$, and the elements are defined by 
\begin{equation}\label{eq3}
\left(\mathcal {X} \circ \mathcal {Y}\right)_{i_1,i_2,\ldots,i_N, i_1',i_2',\ldots,i_M'}\ =\ x_{i_1,i_2,\cdots,i_N} y_{i_1',i_2',\cdots,i_M'}
\end{equation}
for all values of the indices.
\end{definition}
Notice that for rank-one tensors $\mathcal{X}=\mathbf{x}^{(1)} \circ \mathbf{x}^{(2)} \circ \cdots \circ \mathbf{x}^{(M)}$ and $\mathcal{Y}=\mathbf{y}^{(1)} \circ \mathbf{y}^{(2)} \circ  \cdots \circ \mathbf{y}^{(M)}$, it holds that 
\begin{equation}
\left\langle \mathcal {X}, \mathcal {Y}\right\rangle=\left\langle \mathbf{x}^{(1)}, \mathbf{y}^{(1)}\right\rangle \left\langle \mathbf{x}^{(2)}, \mathbf{y}^{(2)}\right\rangle \cdots \left\langle \mathbf{x}^{(M)}, \mathbf{y}^{(M)}\right\rangle.
\label{eq:inner_rank1}
\end{equation}



\begin{definition}[CP factorization~\cite{kolda2009tensor}]
Given a tensor $\mathcal{X} \in \mathbb{R}^{I_{1} \times I_{2} \times \cdots \times I_{M}}$ and an integer $R$, the CP factorization is defined by factor matrices $\mathbf{X}^{(m)} \in \mathbb{R}^{I_m \times R}$ for $m \in [1:M]$, respectively, such that 
\begin{align}
\mathcal{X} = \sum_{r=1}^{R} \mathbf{x}_{r}^{(1)} \circ \mathbf{x}_{r}^{(2)} \circ \cdots  \circ \mathbf{x}_{r}^{(M)} = \llbracket \mathbf{X}^{(1)}, \mathbf{X}^{(2)}, \cdots, \mathbf{X}^{(M)} \rrbracket ~,
\end{align}
where $\mathbf{x}_{r}^{(m)} \in \mathbb{R}^{I_m}$ is the $r$-th column of the factor matrix $\mathbf{X}^{(m)}$, and $\llbracket \cdot \rrbracket$ is used for shorthand notation of the sum of rank-one tensors.
\end{definition}

\subsection{Problem Formulation}\label{sec:problem}
Our problem is different from conventional multi-view learning approaches where multiple views of data are assumed independent and disjoint, and each view is described by a vector. We formulate the multi-view learning problem using coupled analysis of multi-view features in the form of multiple tensors. 

%


Suppose that the problem includes $V$ views where each view consists of a collection of subsets of entities (such as person, company, location, product) and different views have some entities in common. We denote a view as a tuple $(\mathbf{x}^{(1)}, \mathbf{x}^{(2)}, \cdots, \mathbf{x}^{(M)}), M \geq 2$, where $\mathbf{x}^{(m)} \in \mathbb{R}^{I_m}$ is a feature vector associated with the entity $m$.
Inspired by \cite{cao2016multi}, we construct tensor representation for each view over its entities by
\[
\tilde{\mathcal{X}} = \tilde{\mathbf{x}}^{(1)} \circ \tilde{\mathbf{x}}^{(2)} \circ \cdots \circ \tilde{\mathbf{x}}^{(M)} \in \mathbb{R}^{(1+I_1) \times \cdots \times (1+I_M)},
\]
where $\tilde{\mathbf{x}}^{(m)} = [1; \mathbf{x}^{(m)}] \in \mathbb{R}^{1+ I_m}$ and $\circ$ is the outer product operator. In this manner, the full-order interactions \footnote{Full-order interactions range from the first-order interactions (i.e., contributions of single entity features) to the highest-order interactions  (i.e., contributions of the outer product of features from all entities).} between entities are embedded within the tensor structure, which not only provides a unified and compact representation for each view, but also facilitate efficient design methods. Fig.~\ref{fig:input} shows an example of two structural views, where the first view consists of the full-order interactions among the first three modes (e.g., review text, item ID, and user ID), and the second view consists of the full-order interactions among the last two modes (e.g., user ID and friend IDs). 

After generating the tensor representation for each view, we define the multi-view learning problem as follows. Given a training set $\mathfrak{D} = \big \{ \big ( \big \{ \tilde{\mathcal{X}}^{(1)}_{n}, \tilde{\mathcal{X}}^{(2)}_{n}, \cdots, \tilde{\mathcal{X}}^{(V)}_{n} \big \}, ~ y_{n} \big )~ |~ n \in [1 : N] \big \}$, where $\tilde{\mathcal{X}}^{(v)}_{n} \in \mathbb{R}^{(1+I_1) \times \cdots \times (1+I_{M_v})}$ is the tensor representation in the $v$-th view for the $n$-th instance, $y_{n}$ is the response of the $n$-th instance, $M_v$ is the number of the constitutive modes in the $v$-th view, and $N$ is the number of labeled instances. 
We assume different views have common entities, thus the resulting tensors will share common modes, e.g., the third mode in Fig~\ref{fig:input}. 
As we are concerned with predicting unknown values of multiple coupled tensors, our goal is to leverage the relational information from all the views to help predict the unlabeled instances, as well as to use the complementary information among different views to improve the performance. Specifically, we are interested in finding a predictive function $f : \mathfrak{X}^{(1)} \times \mathfrak{X}^{(2)} \cdots \times \mathfrak{X}^{(V)} \rightarrow \mathfrak{Y}$ that minimizes the expected loss, where $\mathfrak{X}^{(v)}, v \in [1 : V] $ is the input space in the $v$-th view and $\mathfrak{Y}$ is the output space.

%% file: table_notation.tex
\begin{table}
\caption{List of basic symbols.}
\label{tab:notation}
\begin{tabular}{ll}
\toprule
Symbol & Definition and description\\
\midrule
$x$ & each lowercase letter represents a scalar\\
$\mathbf{x}$ & each boldface lowercase letter represents a vector\\
$\mathbf{X}$ & each boldface uppercase letter represents a matrix\\
$\mathcal{X}$ & each calligraphic letter represents a tensor \\
$\mathfrak{X}$ & each gothic letter represent a general set or space \\
$[1:N]$ & a set of integers in the range of $1$ to $N$ inclusively. \\
$\left\langle \cdot, \cdot \right\rangle$ & denotes inner product\\
$\circ$ & denotes tensor product (outer product)\\
$\ast$ & denotes Hadamard (element-wise) product\\
\bottomrule
\end{tabular}
\end{table}

%% file: 03-model.tex
\section{Methodology}\label{sec:model}
In this section, we first discuss how to design the predictive models for learning from multiple coupled tensors. We then derive structural factorization machines (SFMs) that can learn the common latent spaces shared in multi-view coupled tensors and automatically adjust the importance of each view in the predictive model.
\subsection{Predictive Models} 
Without loss of generality, we take two views as an example to introduce our basic design of the predictive models. Specifically, we consider coupled analysis of a third-order tensor and a matrix with one mode in common, as shown in Fig.~\ref{fig:input}. Given an input instance $\big ( \big\{ \tilde{\mathcal{X}}^{(1)}, \tilde{\mathbf{X}}^{(2)} \big \}, ~y \big)$, where $\tilde{\mathcal{X}}^{(1)} = \tilde{\mathbf{x}}^{(1)} \circ \tilde{\mathbf{x}}^{(2)} \circ \tilde{\mathbf{x}}^{(3)} \in \mathbb{R}^{(1+I) \times (1+J) \times (1+K)}$ and $\tilde{\mathbf{X}}^{(2)} = \tilde{\mathbf{x}}^{(3)} \circ \tilde{\mathbf{x}}^{(4)} \in \mathbb{R}^{(1+K) \times (1+L)}$. An intuitive solution is to build the following multiple linear model:
\begin{equation}
f \left ( \left \{ \tilde{\mathcal{X}}^{(1)}, \tilde{\mathbf{X}}^{(2)} \right \} \right) = \left \langle \tilde{\mathcal{W}}^{(1)}, \tilde{\mathcal{X}}^{(1)} \right \rangle + \left \langle \tilde{\mathbf{W}}^{(2)}, \tilde{\mathbf{X}}^{(2)} \right \rangle
\label{eq:sfm_base}
\end{equation}
where $\tilde{\mathcal{W}}^{(1)} \in \mathbb{R}^{(1+I) \times (1+J) \times (1+K)}$ and $\tilde{\mathbf{W}}^{(2)} \in \mathbb{R}^{ (1+K) \times (1+L)}$ are the weights for each view to be learned.

\begin{figure*}[t]
\centering
      \includegraphics[width=1.8\columnwidth]{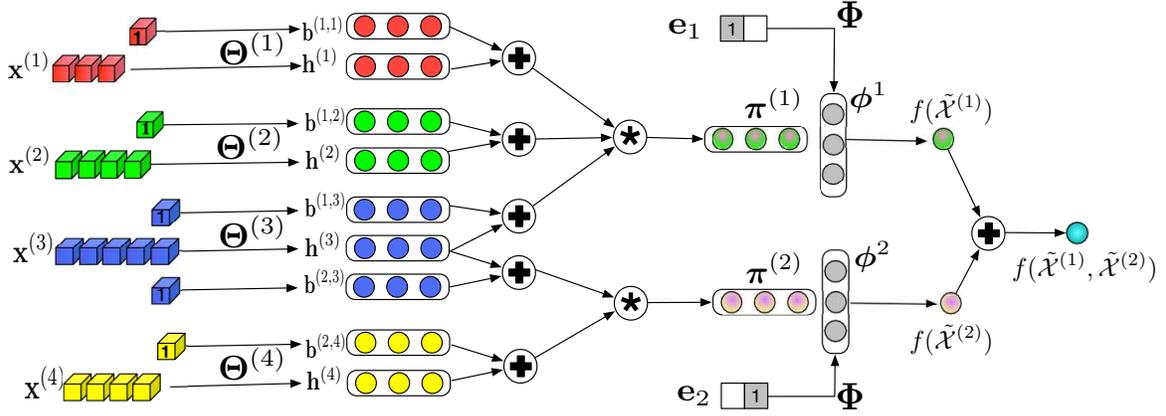}
  \caption{Example of the computational graph in a structural factorization machine, given the input $\tilde{\mathcal{X}}^{(1)}$ and $\tilde{\mathbf{X}}^{(2)}$. By jointly factorizing weight tensors, the $\mathbf{h}^{(m)}$ can be regarded as the latent representation of the feature $\mathbf{x}^{(m)}$ in $m$-th mode, and $\bm{\pi}^{(v)}$ can be regarded as the joint representation of all the modes in the $v$-th view, which can be easily computed through the Hadamard product. The contribution of $\bm{\pi}^{(v)}$ to the final prediction score is automatically adjusted by the weight vector $\bm{\phi}^v$. 
  }\label{fig:sfm}
\end{figure*}

However, in this case it does not take into account the relations and differences between two views. 
In order to incorporate the relations between two views and also discriminate the importance of each view, we introduce an indicator vector $\mathbf{e}_v \in \mathbb{R}^{V}$ for each view $v$ as
\[
\mathbf{e}_v = [ \underbrace{0, \cdots, 0}_\text{v-1}, 1, 0, \cdots, 0 ]^\mathrm{T},
\]
and transform the predictive model in Eq.~(\ref{eq:sfm_base}) into
\begin{equation}
f \left ( \left \{ \tilde{\mathcal{X}}^{(1)}, \tilde{\mathbf{X}}^{(2)} \right \} \right) = \left \langle \hat{\mathcal{W}}^{(1)}, \tilde{\mathcal{X}}^{(1)} \circ \mathbf{e}_1 \right \rangle + \left \langle \hat{\mathcal{W}}^{(2)}, \tilde{\mathbf{X}}^{(2)} \circ \mathbf{e}_2 \right \rangle, 
\label{eq:sfm_view}
\end{equation}
where $\hat{\mathcal{W}}^{(1)} \in \mathbb{R}^{(1+I) \times (1+J) \times (1+K) \times 2}$ and $\hat{\mathcal{W}}^{(2)} \in \mathbb{R}^{(1+K) \times (1+L) \times 2}$. 

Directly learning the weight tensors $\hat{\mathcal{W}}$s leads to two drawbacks. 
First, the weight parameters are learned independently for different modes and different views. When the feature interactions rarely (or even never) appear during training, it is unlikely to learn the associated parameters appropriately. 
Second, the number of parameters in Eq.~(\ref{eq:sfm_view}) is exponential to the number of features, which can make the model prone to overfitting and ineffective on sparse data. 
Here, we assume that each weight tensor has a low-rank approximation, and $\hat{\mathcal{W}}^{(1)}$ and $\hat{\mathcal{W}}^{(2)}$ can be decomposed by CP factorization as 
\begin{align*}
\hat{\mathcal{W}}^{(1)} &= \llbracket \hat{\mathbf{\Theta}}^{(1,1)}, \hat{\mathbf{\Theta}}^{(1, 2)}, \hat{\mathbf{\Theta}}^{(1,3)}, \mathbf{\Phi} \rrbracket \nonumber \\
 &= \llbracket [\mathbf{b}^{(1,1)}; \mathbf{\Theta}^{(1)}], [\mathbf{b}^{(1,2)}; \mathbf{\Theta}^{(2)}], [\mathbf{b}^{(1,3)}; \mathbf{\Theta}^{(3)}], \mathbf{\Phi} \rrbracket ,
\end{align*}
 and 
\begin{align*}
\hat{\mathcal{W}}^{(2)} = \llbracket \hat{\mathbf{\Theta}}^{(2, 3)}, \hat{\mathbf{\Theta}}^{(2, 4)}, \mathbf{\Phi} \rrbracket 
 = \llbracket [\mathbf{b}^{(2,3)}; \mathbf{\Theta}^{(3)}], [\mathbf{b}^{(2,4)}; \mathbf{\Theta}^{(4)}], \mathbf{\Phi} \rrbracket, 
\end{align*}
where $\bm{\Theta}^{(m)} \in \mathbb{R}^{I_m \times R}$ is the factor matrix for the features in the $m$-th mode. It is worth noting that $\mathbf{\Theta}^{(3)}$ is shared in the two views. 
$\bm{\Phi} \in \mathbb{R}^{2 \times R}$ is the factor matrix for the view indicator, and $\mathbf{b}^{(v,m)} \in \mathbb{R}^{1 \times R}$, which is always associated with the constant one in $\tilde{\mathbf{x}}^{(m)} = [1;\mathbf{x}^{(m)}]$, represents the bias factors of the $m$-th mode in the $v$-th view. Through $\mathbf{b}^{(v,m)}$, the lower-order interactions (the interactions excluding the features from the $m$-th mode) in the $v$-th view are explored in the predictive function. 

Then we can transform Eq.~(\ref{eq:sfm_view}) into
\begin{equation}
\begin{adjustbox}{max width=1\columnwidth}
$\displaystyle
\begin{aligned}
&\left \langle \hat{\mathcal{W}}^{(1)}, \tilde{\mathcal{X}}^{(1)} \circ \mathbf{e}_1 \right \rangle + \left \langle \hat{\mathcal{W}}^{(2)}, \tilde{\mathbf{X}}^{(2)} \circ \mathbf{e}_2 \right \rangle \\
= & \sum_{r=1}^{R}  \left\langle  \hat{\bm{\theta}}_r^{(1,1)} \circ \hat{\bm{\theta}}_r^{(1,2)}  \circ \hat{\bm{\theta}}_r^{(1,3)} \circ \bm{\phi}_{r} ~,~ \tilde{\mathbf{x}}^{(1)} \circ \tilde{\mathbf{x}}^{(2)} \circ \tilde{\mathbf{x}}^{(3)} \circ \mathbf{e}_1 \right\rangle \\
& + \sum_{r=1}^{R}  \left\langle \hat{\bm{\theta}}_r^{(2,3)} \circ \hat{\bm{\theta}}_r^{(2,4)} \circ \bm{\phi}_{r} ~,~ \tilde{\mathbf{x}}^{(3)} \circ \tilde{\mathbf{x}}^{(4)} \circ \mathbf{e}_2 \right\rangle \\
= & \bm{\phi}^1 \left ( \prod_{m=1}^{3} \ast \left( \tilde{\mathbf{x}}^{(m)^\mathrm{T}} \hat{\bm{\Theta}}^{(1,m)} \right)  \right)^\mathrm{T}
+ \bm{\phi}^2 \left ( \prod_{m=3}^{4} \ast \left( \tilde{\mathbf{x}}^{(m)^\mathrm{T}} \hat{\bm{\Theta}}^{(2,m)} \right) \right)^\mathrm{T} \\
= & \bm{\phi}^1 \left ( \prod_{m=1}^{3} \ast \left( \mathbf{x}^{(m)^\mathrm{T}} \bm{\Theta}^{(m)} + \mathbf{b}^{(1,m)} \right)  \right)^\mathrm{T}
+ \bm{\phi}^2 \left ( \prod_{m=3}^{4} \ast \left( \mathbf{x}^{(m)^\mathrm{T}} \bm{\Theta}^{(m)} + \mathbf{b}^{(2,m)} \right) \right)^\mathrm{T}
\label{eq:sfm_cp}
\end{aligned}
$
\end{adjustbox}
\end{equation}

where $\ast$ is the Hadamard (elementwise) product and $ \bm{\phi}^v \in \mathbb{R}^{1 \times R}$ is the $v$-th row of the factor matrix $\bm{\Phi}$. 

For convenience, we let $\mathbf{h}^{(m)} = \bm{\Theta}^{(m)^\mathrm{T}}\mathbf{x}^{(m)}$, $S_M(v)$ denote the set of modes in the $v$-th views, $\bm{\pi}^{(v)} = \prod\limits_{m \in S_M(v)} \ast \left( \mathbf{h}^{(m)} + \mathbf{b}^{(v,m)^\mathrm{T}} \right)$, and $\bm{\pi}^{(v,-m)} = \prod\limits_{m' \in S_M(v), m' \neq m}$ $\ast \left( \mathbf{h}^{(m')} + \mathbf{b}^{(v,m')^\mathrm{T}} \right)$. 
The predictive model for the general cases is given as follows 
\begin{equation}
\begin{adjustbox}{max width=0.95\columnwidth}
$\displaystyle
\begin{aligned}
f(\{\tilde{\mathcal{X}}^{(v)}\}) &= \sum_{v=1}^{V} \left\langle \hat{\mathcal{W}}^{(v)},\tilde{\mathcal{X}}^{(v)} \circ \mathbf{e}_v \right\rangle  \\
& = \sum_{v=1}^{V} \bm{\phi}^v \prod_{m \in S_M(v)} \ast \left( \mathbf{x}^{(m)^\mathrm{T}}\bm{\Theta}^{(m)}  + \mathbf{b}^{(v,m)} \right)^\mathrm{T} \\
& = \sum_{v=1}^{V} \bm{\phi}^v \prod_{m \in S_M(v)} \ast \left( \mathbf{h}^{(m)} + \mathbf{b}^{(v,m)^\mathrm{T}} \right)
\label{eq:mfm_general}
\end{aligned}
$
\end{adjustbox}
\end{equation}
A graphical illustration of the proposed model is shown in Fig.~\ref{fig:sfm}. We name this model as structural factorization machines (SFMs). 
Clearly, the parameters are jointly factorized, which benefits parameter estimation under sparsity since dependencies exist when the interactions share the same features. Therefore, the model parameters can be effectively learned without direct observations of such interactions especially in highly sparse data. 
More importantly, after factorizing the weight tensor $\hat{\mathcal{W}}$s, there is no need to construct the input tensor physically. 
Furthermore, the model complexity is linear in the number of original features. In particular, the model complexity is $O(R(V+I + \sum_{v}M_v))$, where $M_v$ is the number of modes in the $v$-th view.  


%% file: 04-learning.tex

\subsection{Learning Structural Factorization Machines}\label{sec:learning}
Following the traditional supervised learning framework, we propose to learn the model parameters by minimizing the following regularized empirical risk:
\begin{equation}
\mathcal{R}= \frac{1}{N} \sum_{n=1}^{N} \ell \left(f(\{\mathcal{X}_{n}^{(v)}\} ), y_{n} \right) + \lambda \Omega( \mathbf{\Phi},\{\mathbf{\Theta}^{(m)}\}, \{\mathbf{b}^{(v,m)}\}) 
\label{eq:empirical_formula}
\end{equation}
where $\ell$ is a prescribed loss function, $\Omega$ is the regularizer encoding the prior knowledge of $\{\mathbf{\Theta}^{(m)} \}$ and $\mathbf{\Phi}$, and $\lambda \geq 0$ is the regularization parameter that controls the trade-off between the empirical loss and the prior knowledge.




The partial derivative of $\mathcal{R}$ w.r.t. $\mathbf{\Theta}^{(m)}$ is given by 
\begin{align}
\frac{\partial \mathcal{R}}{\partial \mathbf{\Theta}^{(m)}} = \frac{\partial \mathcal{L}}{\partial f} \frac{\partial f}{\partial \mathbf{\Theta}^{(m)}} + \lambda  \frac{\partial \Omega_{\lambda}( \mathbf{\Theta}^{(m)}) }{\partial \mathbf{\Theta}^{(m)}}
\label{eq:mfm_grad_theta}
\end{align}
where $\frac{\partial \mathcal{L}}{\partial f} = \frac{1}{N} 
\left[\begin{array}{c} 
\frac{\partial \ell_{1} }{\partial f}, \cdots, \frac{\partial \ell_{N} }{\partial f}
\end{array}\right]^\mathrm{T} \in \mathbb{R}^{N}$. 

For convenience, we let $S_V(m)$ denote the set of views that contains the $m$-th mode, $\mathbf{X}^{(m)} = [\mathbf{x}_{1}^{(m)}, \cdots, \mathbf{x}_{N}^{(m)}]$, $\bm{\Pi}^{(v)} = [\bm{\pi}_{1}^{(v)}, \cdots, \bm{\pi}_{N}^{(v)}]^\mathrm{T}$ and $\bm{\Pi}^{(v,-m)} = [\bm{\pi}_{1}^{(v,-m)}, \cdots, \bm{\pi}_{N}^{(v,-m)}]^\mathrm{T}$. We then have that 
\begin{equation}
 \frac{\partial \mathcal{L}}{\partial f} \frac{\partial f}{\partial \mathbf{\Theta}^{(m)}} 
= \mathbf{X}^{(m)}\left( \sum_{v \in S_V(m)}\left( \left( \frac{\partial \mathcal{L}}{\partial f} \bm{\phi}^{v} \right) \ast \bm{\Pi}^{(v,-m)} \right) \right)
\label{eq:mfm_grad_Lt_theta}
\end{equation}

Similarly, the partial derivative of $\mathcal{R}$ w.r.t. $\mathbf{b}^{(v,m)}$ is given by 
\begin{align}
\frac{\partial \mathcal{R}}{\partial \mathbf{b}^{(v,m)}} &= \frac{\partial \mathcal{L}}{\partial f} \frac{\partial f}{\partial \mathbf{b}^{(v,m)}} + \lambda  \frac{\partial \Omega_{\lambda}( \mathbf{b}^{(v,m)}) }{\partial \mathbf{b}^{(v,m)}}\nonumber \\
&= \mathbf{1}^\mathrm{T} \left( \left( \frac{\partial \mathcal{L}}{\partial f} \bm{\phi}^{v} \right) \ast \bm{\Pi}^{(v,-m)} \right)+ \lambda  \frac{\partial \Omega_{\lambda}( \mathbf{b}^{(v,m)}) }{\partial \mathbf{b}^{(v,m)}}
\label{eq:mfm_grad_b}
\end{align}


The partial derivative of $\mathcal{R}$ w.r.t. $\mathbf{\Phi}$ is given by 
\begin{align}
\frac{\partial \mathcal{R}}{\partial \mathbf{\Phi}} 
&=
\left[\begin{array}{c} 
 \left( \frac{\partial \mathcal{L}}{\partial f} \right)^{\mathrm{T}} \bm{\Pi}^{(1)}~; ~\cdots ;~\left( \frac{\partial \mathcal{L}}{\partial f} \right)^{\mathrm{T}} \bm{\Pi}^{(V)}
\end{array}\right]
+ \lambda  \frac{\partial \Omega_{\lambda}( \mathbf{\Phi})}{\partial \mathbf{\Phi}}
\label{eq:mfm_grad_phi}
\end{align}


Finally, the gradient of $\mathcal{R}$ can be formed by vectorizing the partial derivatives with respect to each factor matrix and concatenating them all, i.e., 
\begin{align}
    \nabla \mathcal{R} = \left[ \begin{array}{c} 
\text{vec}(\frac{\partial \mathcal{R}}{\partial \bm{\Theta}^{(1)}})\\ 
\vdots\\
\text{vec}(\frac{\partial \mathcal{R}}{\partial \bm{\Theta}^{(M)}})\\
\text{vec}(\frac{\partial \mathcal{R}}{\partial \mathbf{b}^{(1,1)}})\\
\vdots\\
\text{vec}(\frac{\partial \mathcal{R}}{\partial \mathbf{b}^{(V,M)}})\\
\text{vec}(\frac{\partial \mathcal{R}}{\partial \bm{\Phi}})
\end{array} \right]
\end{align}



Once we have the function, $\mathcal{R}$ and gradient, $\nabla \mathcal{R}$, we can use any gradient-based optimization algorithm to compute the factor matrices. For the results presented in this paper, we use the Adaptive Moment Estimation (Adam) optimization algorithm~\cite{kingma2014adam} for parameter updates. Adam is an adaptive version of gradient descent that controls individual adaptive learning rates for different parameters from estimates of first and second moments of the gradient. It combines the best properties of the AdaGrad~\cite{duchi2011adaptive}, which works well with sparse gradients, and RMSProp~\cite{hinton2012rmsprop}, which works well in on-line and non-stationary settings. Readers can refer to~\cite{kingma2014adam} for details of the Adam optimization algorithm. 

\subsection{Efficient Computing with Relational Structures}\label{sec:computing}
\begin{figure}[t]
\centering
      \includegraphics[width=\columnwidth]{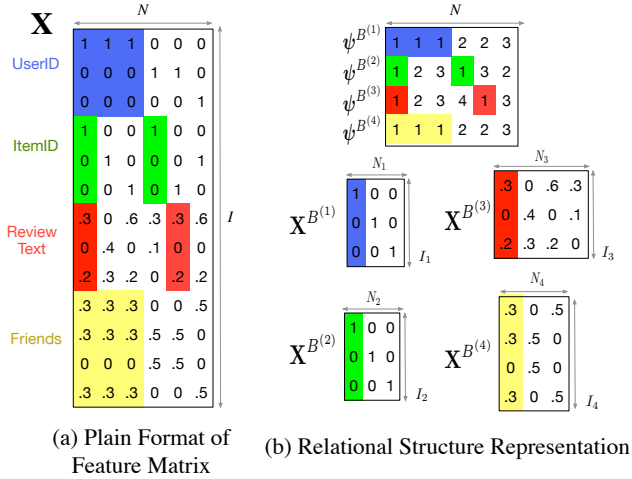}
  \caption{(a) Feature vectors of the same entity repeatedly appear in the plain formatted feature matrix $\mathbf{X}$. 
  (b) Repeating patterns in $\mathbf{X}$ can be formalized by the relational structure $\mathbf{B}$ of each mode.
  For example, the forth column of the feature matrix $\mathbf{X}$ can be represented as $\mathbf{x}_4 = [\mathbf{x}_{\psi(4)}^{(1)}; \mathbf{x}_{\psi(4)}^{(2)}; \mathbf{x}_{\psi(4)}^{(3)}; \mathbf{x}_{\psi(4)}^{(4)}]$ $=[\mathbf{x}_2^{B^{(1)}}; \mathbf{x}_1^{B^{(2)}}; \mathbf{x}_4^{B^{(3)}}; \mathbf{x}_2^{B^{(4)}}]$.
  }\label{fig:block}
\end{figure}


In relational domains, we can often observe that feature vectors of the same entity repeatedly appear in the plain formatted feature matrix $\mathbf{X}$, where $\mathbf{X} = [\mathbf{X}^{(1)};\cdots;\mathbf{X}^{(M)}] \in \mathbb{R}^{I \times N}$ and $\mathbf{X}^{(m)} \in \mathbb{R}^{I_m \times N}$ is the feature matrix in the $m$-th mode.  
Consider Fig.~\ref{fig:block}(a) as an example, where the parts highlighted in yellow in the forth mode (which represents the friends of the user) are repeatedly appear in the first three columns. Clearly, these repeating patterns stem from the relational structure of the same entity. 

In the following, we show how the proposed SFM method can make use of relational structure of each mode, 
such that the learning and prediction can be scaled to predictor variables generated from relational data involving relations of high cardinality. 
We adopt the idea from~\cite{rendle2013scaling} to avoid redundant computing on repeating patterns over a set of feature vectors. 

Let $\mathcal{B} = \{(\mathbf{X}^{B^{(m)}}, \psi^{B^{(m)}})\}_{m=1}^{M}$ be the set of relational structures, where $\mathbf{X}^{B^{(m)}} \in \mathbb{R}^{I_m \times N_m}$ denotes the relational matrix of $m$-th mode, $\psi^{B^{(m)}}: \{1, \cdots, N\} \rightarrow \{1,\cdots, N_{m}\}$ denotes the mapping from columns in the feature matrix $\mathbf{X}$ to columns within $\mathbf{X}^{B^{(m)}}$. To shorten notation, the index $B$ is dropped from the mapping $\psi^{B}$ whenever it is clear which block the mapping belongs to. 
From $\mathcal{B}$, one can reconstruct $\mathbf{X}$ by concatenating the corresponding columns of the relational matrices using the mappings. 
For instance, the feature vector $\mathbf{x}_n$ of the $n$-th case in the plain feature matrix $\mathbf{X}$ is represented as
$\mathbf{x}_n = [ \mathbf{x}_{\psi(n)}^{(1)}; \cdots; \mathbf{x}_{\psi(n)}^{(M)}]$. 
Fig.~\ref{fig:block}(b) shows an example how the feature matrix can be represented in relational structures. 
Let $N_z(\mathbf{A})$ denote the number of non-zeros in a matrix $\mathbf{A}$. The space required for using relational structures to represent the input data is $|\mathcal{B}| = NM + \sum_m N_z(\mathbf{X}^{B^{(m)}})$, which is much smaller than $N_z(\mathbf{X})$ if there are repeating patterns in the feature matrix $\mathbf{X}$. 

Now we can rewrite the predictive model in Eq.~(\ref{eq:mfm_general}) as follows
\begin{equation}\label{eq:mfm_relational}
f(\{\mathcal{X}_{n}^{(v)}\} = \sum_{v=1}^{V} \bm{\phi}^v \prod_{m \in S_M(v)} \ast \left( \mathbf{h}_{\psi(n)}^{B^{(m)}} + \mathbf{b}^{(v,m)^\mathrm{T}} \right),
\end{equation}
with the caches $\mathbf{H}^{B^{(m)}} = [\mathbf{h}^{B^{(m)}}_1, \cdots, \mathbf{h}^{B^{(m)}}_{N_m}]$ for each mode, where $\mathbf{h}^{B^{(m)}}_j = \bm{\Theta}^{(m)^\mathrm{T}} \mathbf{x}_j^{B^{(m)}}, ~\forall j \in [1:N_m]$. 

This directly shows how $N$ samples can be efficiently predicted: (i) compute $\mathbf{H}^{B^{(m)}}$ in $O(R N_z(\mathbf{X}^{B^{(m)}}))$ for each mode, (ii) compute $N$ predictions with Eq.~(\ref{eq:mfm_relational}) using caches in $O(R N (V + \sum_v M_v))$. 
With the help of relational structures, SFMs can learn the same parameters and make the same predictions but with a much lower runtime complexity. 

%% file: 05-exp.tex
\section{Experiments}\label{sec:exp}
\input{table_dataset}
\begin{figure}
\centering
      \includegraphics[width=1\columnwidth]{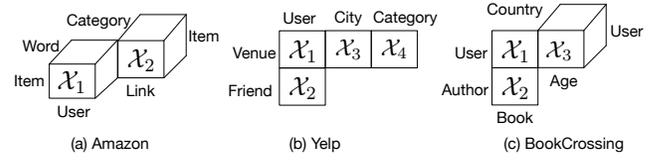}
  \caption{Schema of the structural views in each dataset.}\label{fig:schema}
\end{figure}
\subsection{Datasets}\label{sec:experiments}
To evaluate the ability and applicability of the proposed SFMs, we include a spectrum of large datasets from different domains. 
The statistics for each dataset is summarized in Table \ref{tab:dataset}, the schema of the structural views in each dataset is presented in Fig.~\ref{fig:schema}, and the details are as follows:

	\textbf{Amazon}\footnote{http://jmcauley.ucsd.edu/data/amazon/}:	
    The first group of datasets are from Amazon.com recently introduced by~\cite{mcauley2015inferring}. This is among the largest datasets available that include review texts and metadata of items. Each top-level category of products on Amazon.com has been constructed as an independent dataset in~\cite{mcauley2015inferring}. In this paper, we take a variety of large categories as listed in Tabel~\ref{tab:dataset}. 

 	Each sample in these datasets has five modes, {\em i.e.}, users, items, review texts, categories, and linkage. 
	The user mode and item mode are represented by one-hot encoding. 
 	The $\ell_2$-normalized TF-IDF vector representation of review text~\footnote{Stemming, lemmatization, removing stop-words and  words with frequency less than 100 times, etc., are handled beforehand.} of the item given by the user is used as the text mode. 
 	The category mode and linkage mode consists of all the categories and all the co-purchasing items of the item, which might be from other categories. 
 	The last two modes are $\ell_1$-normalized. 

	\textbf{Yelp}\footnote{https://www.yelp.com/dataset-challenge}:
	It is a large-scale dataset consisting of venue reviews. Each sample in this dataset contains five modes, {\em i.e.}, users, venues, friends, categories and cities. The user mode and venue mode are represented by one-hot encoding. 
	The friend mode consists of the friends' ids of users. 
	The category mode and city mode consists of all the categories and the city of the venue. 
	The last three modes are $\ell_1$-normalized. 
	
	\textbf{BookCrossing (BX)}\footnote{http://www2.informatik.uni-freiburg.de/$\sim$cziegler/BX/}:
	It is a book review dataset collected from the Book-Crossing community. Each sample in this dataset contains five modes, {\em i.e.}, users, books, countries, ages and authors. The ages are split in eight bins as in \cite{harper2016movielens}.	
	The country mode and age mode consist of the corresponding meta information of the user. 
	The author modes represents the authors of the book. 
	All the modes are represented by one-hot encoding. 	
	
	The values of samples range within [1:5] in Amazon and Yelp datasets, and range within [1:10] in BX dataset. 

\subsection{Comparison Methods}\label{sec:compare}
In order to demonstrate the effectiveness of the proposed SFMs, we compare a series of state-of-the-art methods. 

\noindent\textbf{Matrix Factorization (MF)} is used to validate that meta information is helpful for improving prediction performance. We use the LIBMF implementation~\cite{chin2016libmf} for comparison in the experiment.  

\noindent\textbf{Factorization Machine (FM)}~\cite{rendle2012factorization} is the state-of-the-art method in recommender systems. 
We compare with its higher-order extension~\cite{blondel2016higher} with up to second-order, and third-order feature interactions, and denote them as FM-2 and FM-3.

\noindent\textbf{Polynomial Network (PolyNet)}~\cite{livni2014computational} is a recently proposed method that utilizes polynomial kernel on all features. We compare the augmented PolyNet (which adds a constant one to the feature vector~\cite{blondelIFU16}) with up to the second-order, and third-order kernel and denote them as PolyNet-2 and PolyNet-3. 

\noindent\textbf{Multi-View Machine (MVM)}~\cite{cao2016multi} is a tensor factorization based method that explores the latent representation embedded in the full-order interactions among all the modes. 

\noindent\textbf{Structural Factorization Machine (SFM)} is the proposed model that learns the common latent spaces shared in multi-way data. 

\input{table_regression}

\subsection{Experimental Settings}\label{sec:metric}
For each dataset, we randomly split $50\%$, $10\%$, and $40\%$ of labeled samples as training set, validation set, and testing set, respectively. Validation sets are used for hyper-parameter tuning for each model. Each of the validation and testing sets does not overlap with any other set so as to ensure the sanity of the experiment. 
For simplicity and fair comparison, in all the comparison methods, the dimension of latent factors $R=20$ and the maximum number of epochs is set as $400$ and we use early stop to obtain the best results for each method. Forbenius norm regularizers are used to avoid overfitting. The regularization hyper-parameter is tuned from $\{10^{-5},~10^{-4},~\cdots,~10^{0}\}$. 

All the methods except MF are implemented in TensorFlow, and the parameters are initialized using scaling variance initializer~\cite{he2015delving}. We tune the scaling factor of initializer $\sigma$ from $\{1, 2, 5, 10, 100\}$ and the learning rate $\eta$ from $\{0.01, 0.1, 1\}$ using the validation sets. In the experiment, we set $\sigma=2$ (default setting in TensorFlow) and $\eta=0.01$ for these methods except MVM. We found that MVM is more sensitive to the configuration, because MVM will element-wisely multiply the latent factors of all the modes which leads to an extremely small value approaching zero. $\sigma=10$ and $\eta=0.1$ yielded the best performance for MVM. 

To investigate the performance of comparison methods,  we adopt mean squared error (MSE) on the test data as the evaluation metrics~\cite{mcauley2013hidden,zheng2017joint}. The smaller value of the metric indicates the better performance. 
Each experiment was repeated for 10 times, and the mean and standard deviation of each metric in each data set were reported. 
All experiments are conducted on a single machine with Intel Xeon $6$-Core CPUs of 2.4 GHz and equipped with a Maxwell Titan X GPU. 

\begin{figure}
\centering
      \includegraphics[width=\columnwidth]{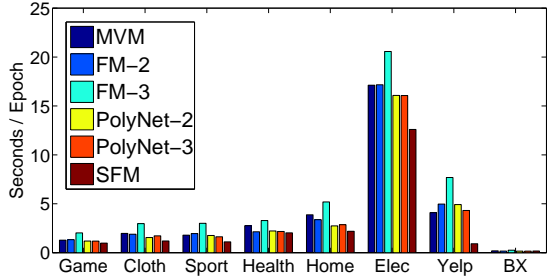}
  \caption{Training Time (Seconds/Epoch) Comparison.}\label{fig:time}
\end{figure}

\begin{figure*}[!t]
\centering
 \subfigure[Sport]{\label{fig:sport_cold}
    \begin{minipage}[l]{0.24\linewidth}
     \centering
     \includegraphics[width=\linewidth]{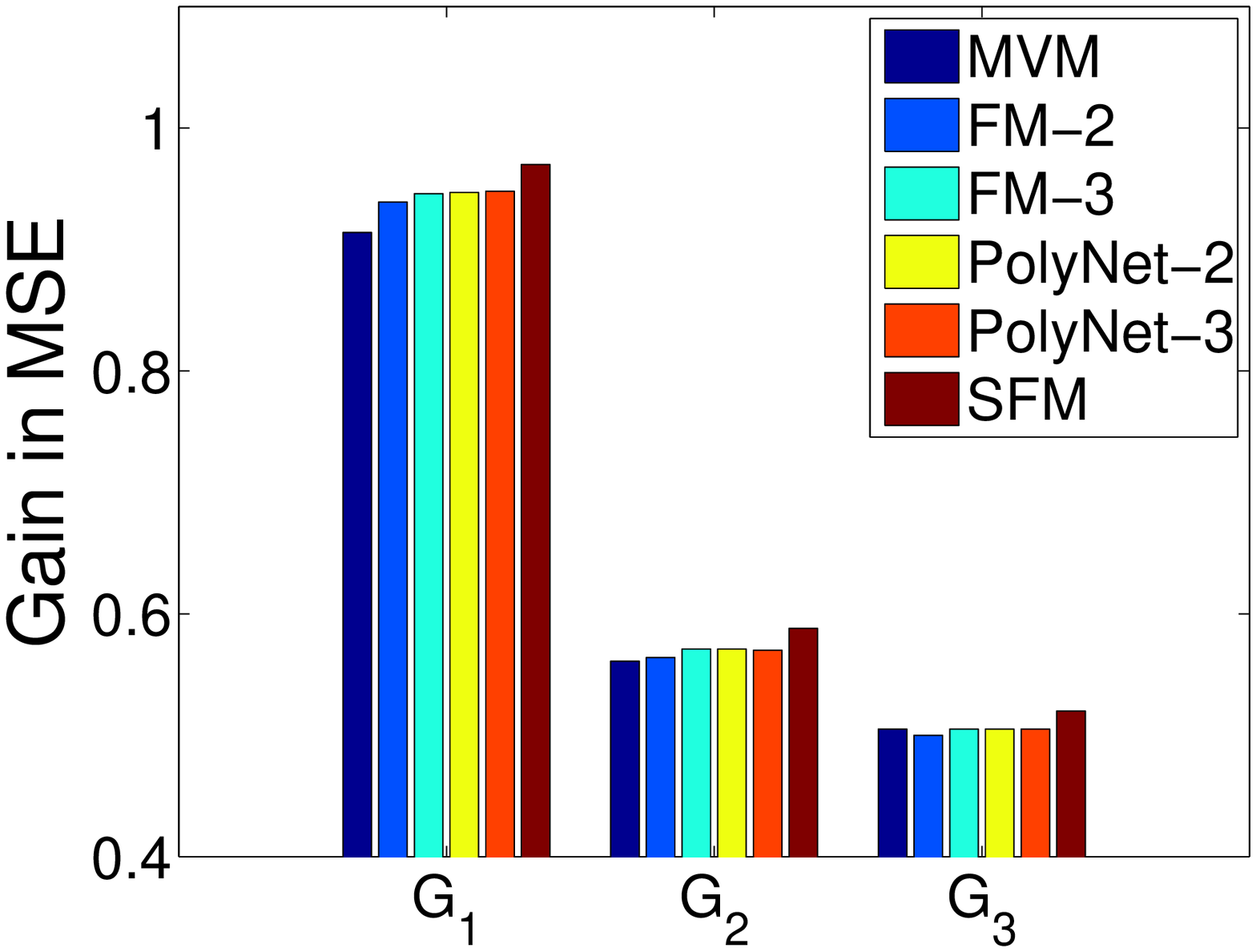}
    \end{minipage}
 }~
 \subfigure[Health]{\label{fig:health_cold}
    \begin{minipage}[l]{0.24\linewidth}
     \centering
     \includegraphics[width=\linewidth]{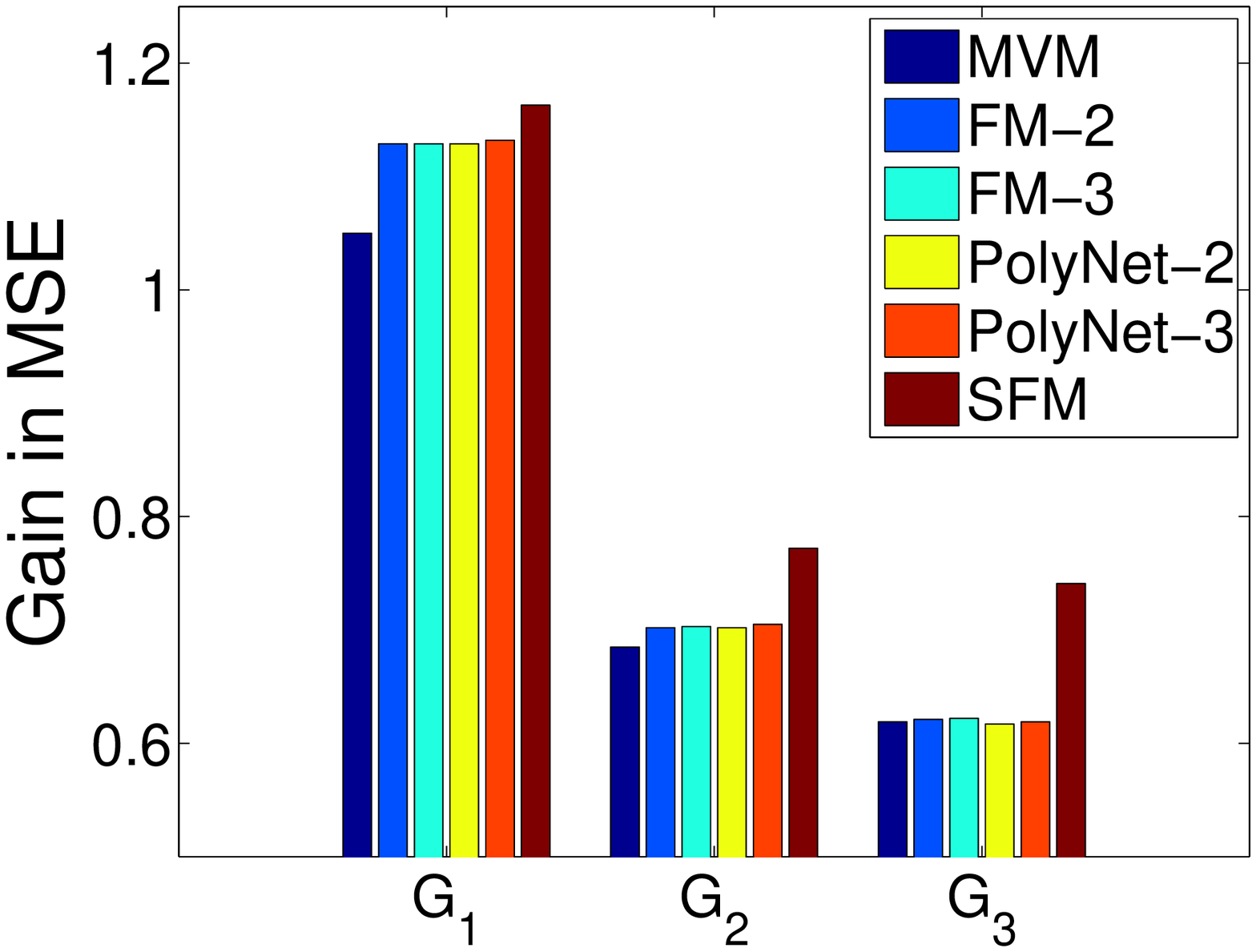}
    \end{minipage}
 }~
\subfigure[Yelp]{\label{fig:yelp_cold}
    \begin{minipage}[l]{0.24\linewidth}
     \centering
     \includegraphics[width=\linewidth]{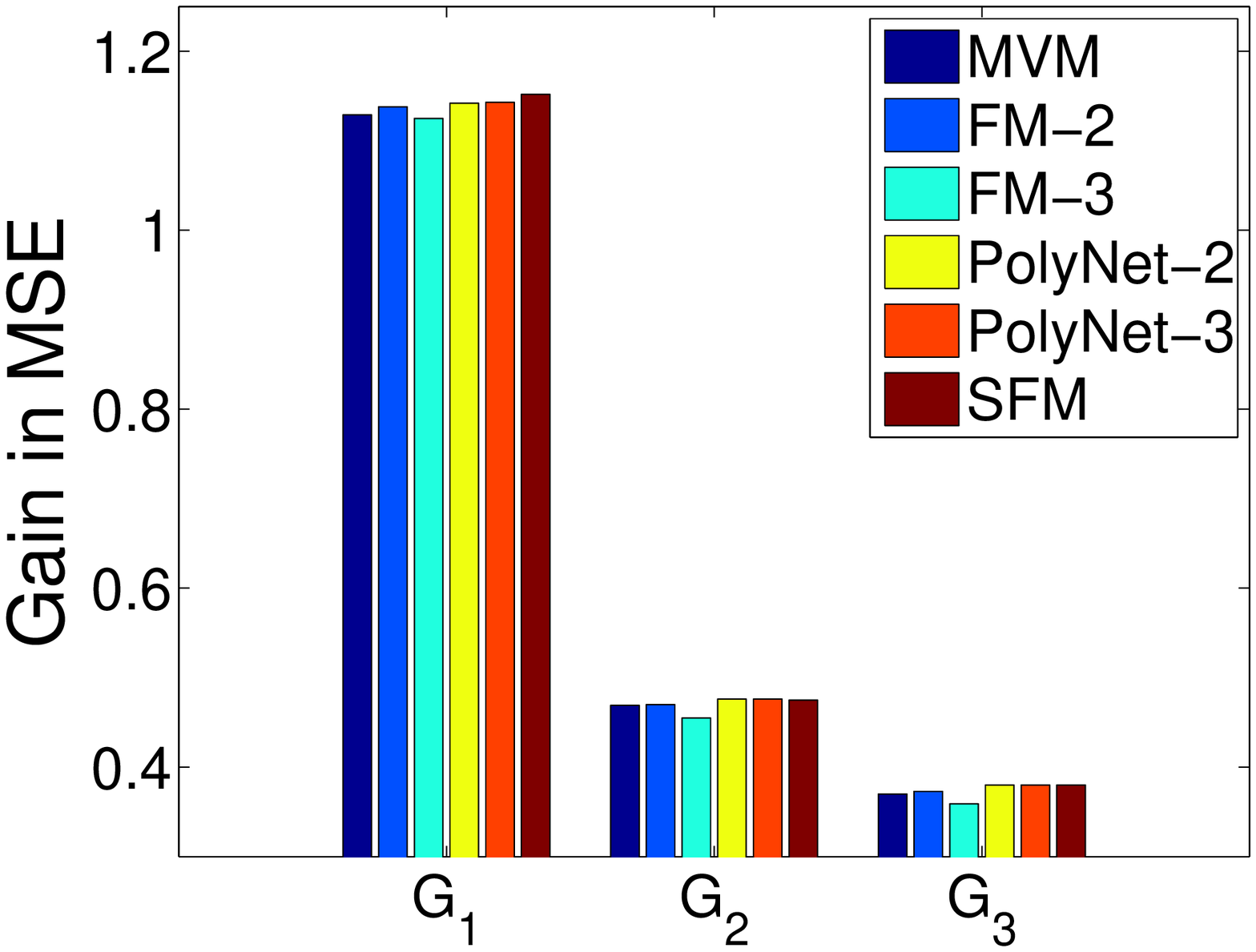}
    \end{minipage}
 }~
 \subfigure[BX]{\label{fig:bx_cold}
    \begin{minipage}[l]{0.24\linewidth}
     \centering
     \includegraphics[width=\linewidth]{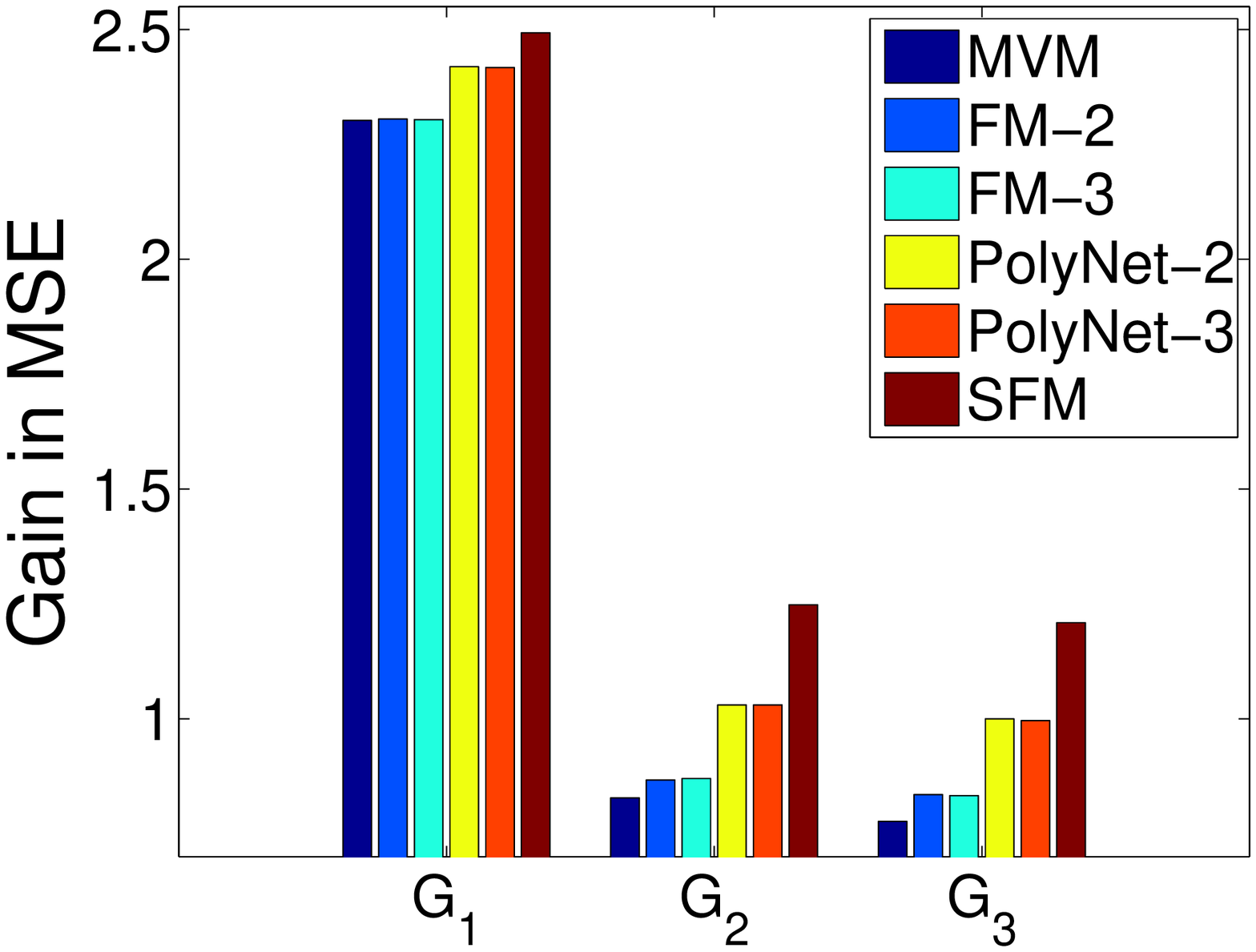}
    \end{minipage}
 }
\caption{Performance gain in MSE compared with MF for users with limited training samples. $G_1$, $G_2$, and $G_3$ are groups of users with $[1,3]$, $[4,6]$, and $[7,10]$ observed samples in the training set, respectively.}\label{fig:exp_cold}
\end{figure*}

\begin{figure*}[!t]
\centering
 \subfigure[Sport]{\label{fig:sport_R}
    \begin{minipage}[l]{0.24\linewidth}
     \centering
     \includegraphics[width=\linewidth]{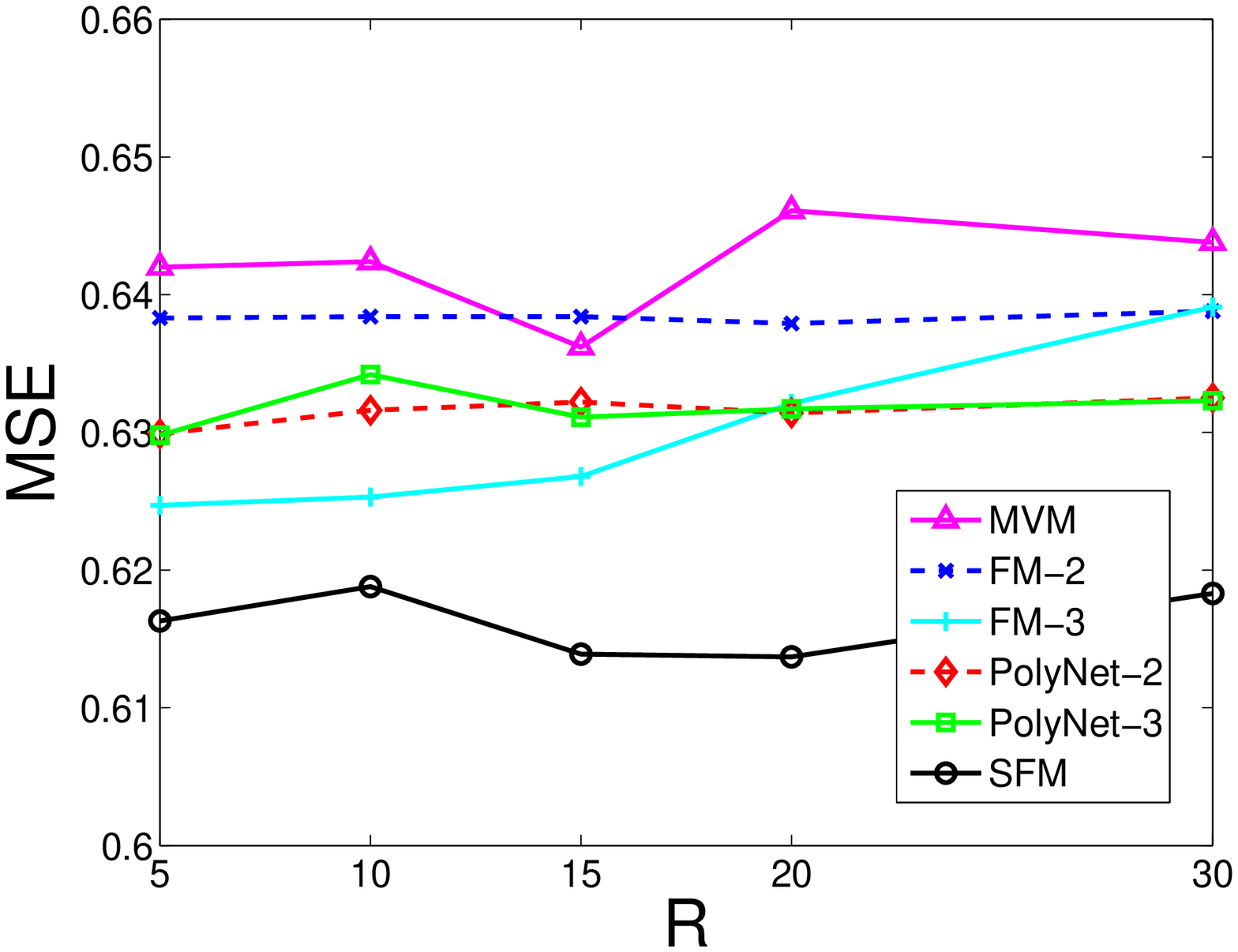}
    \end{minipage}
 }~
 \subfigure[Health]{\label{fig:health_R}
    \begin{minipage}[l]{0.24\linewidth}
     \centering
     \includegraphics[width=\linewidth]{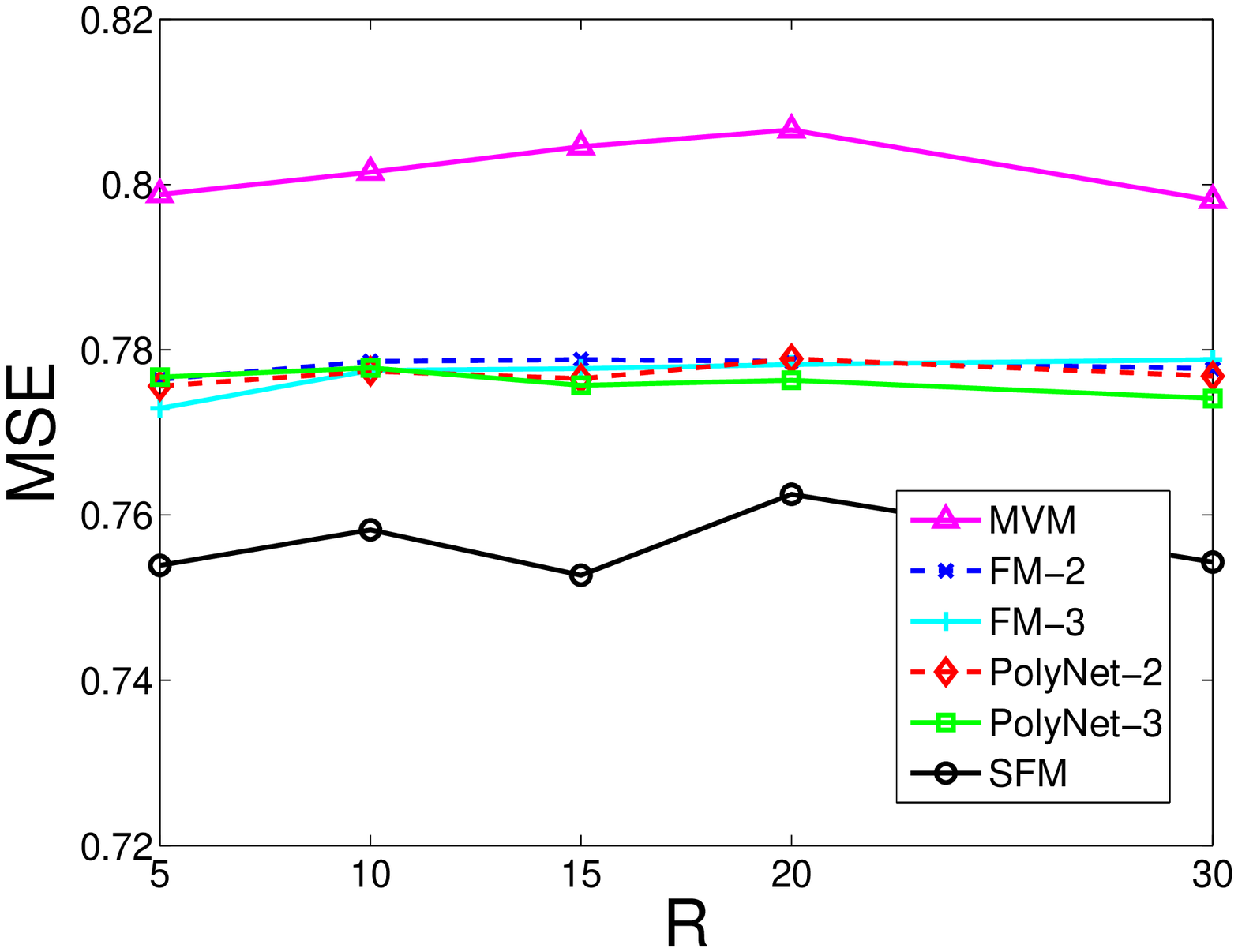}
    \end{minipage}
 }~
\subfigure[Yelp]{\label{fig:yelp_R}
    \begin{minipage}[l]{0.24\linewidth}
     \centering
     \includegraphics[width=\linewidth]{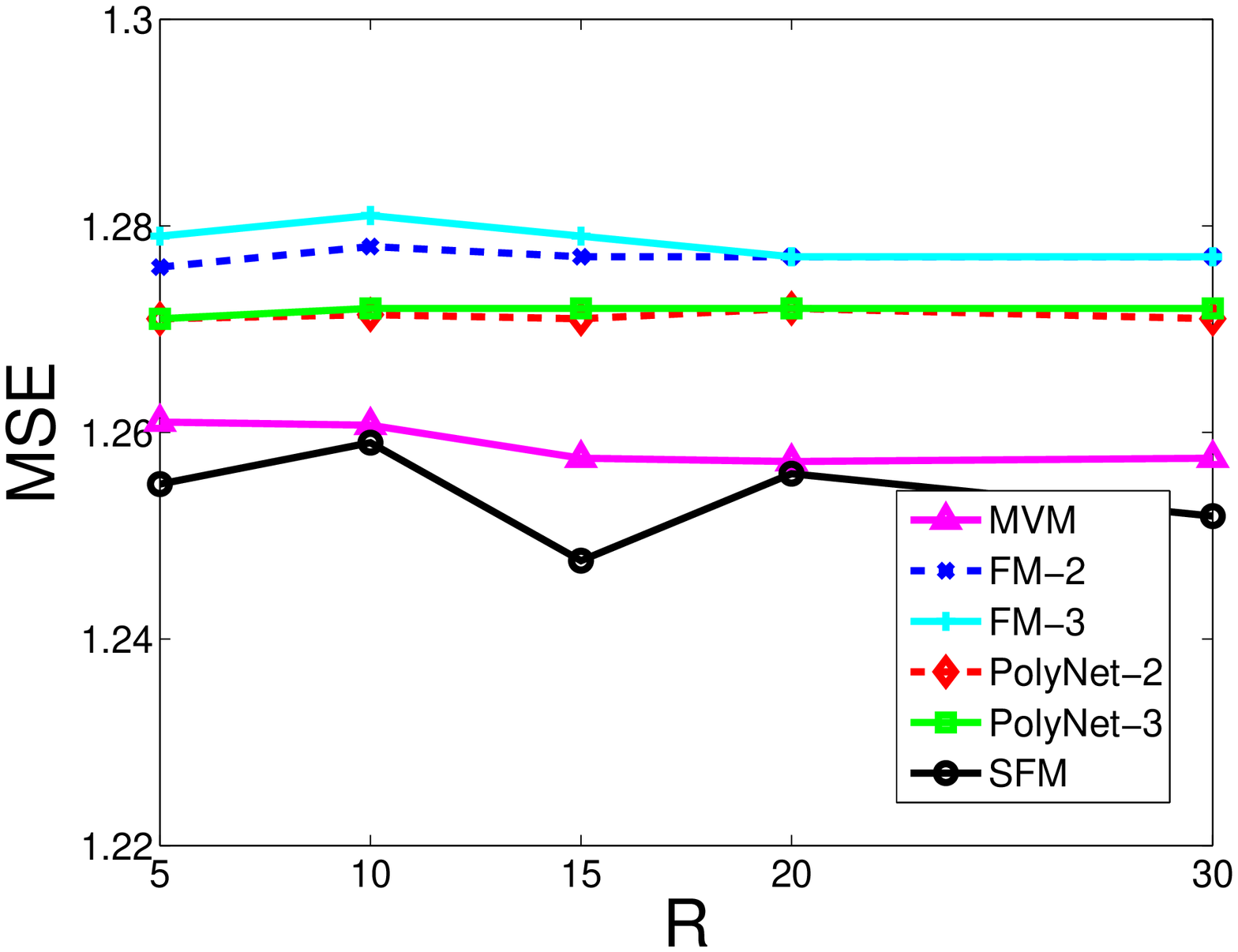}
    \end{minipage}
 }~
 \subfigure[BX]{\label{fig:bx_R}
    \begin{minipage}[l]{0.24\linewidth}
     \centering
     \includegraphics[width=\linewidth]{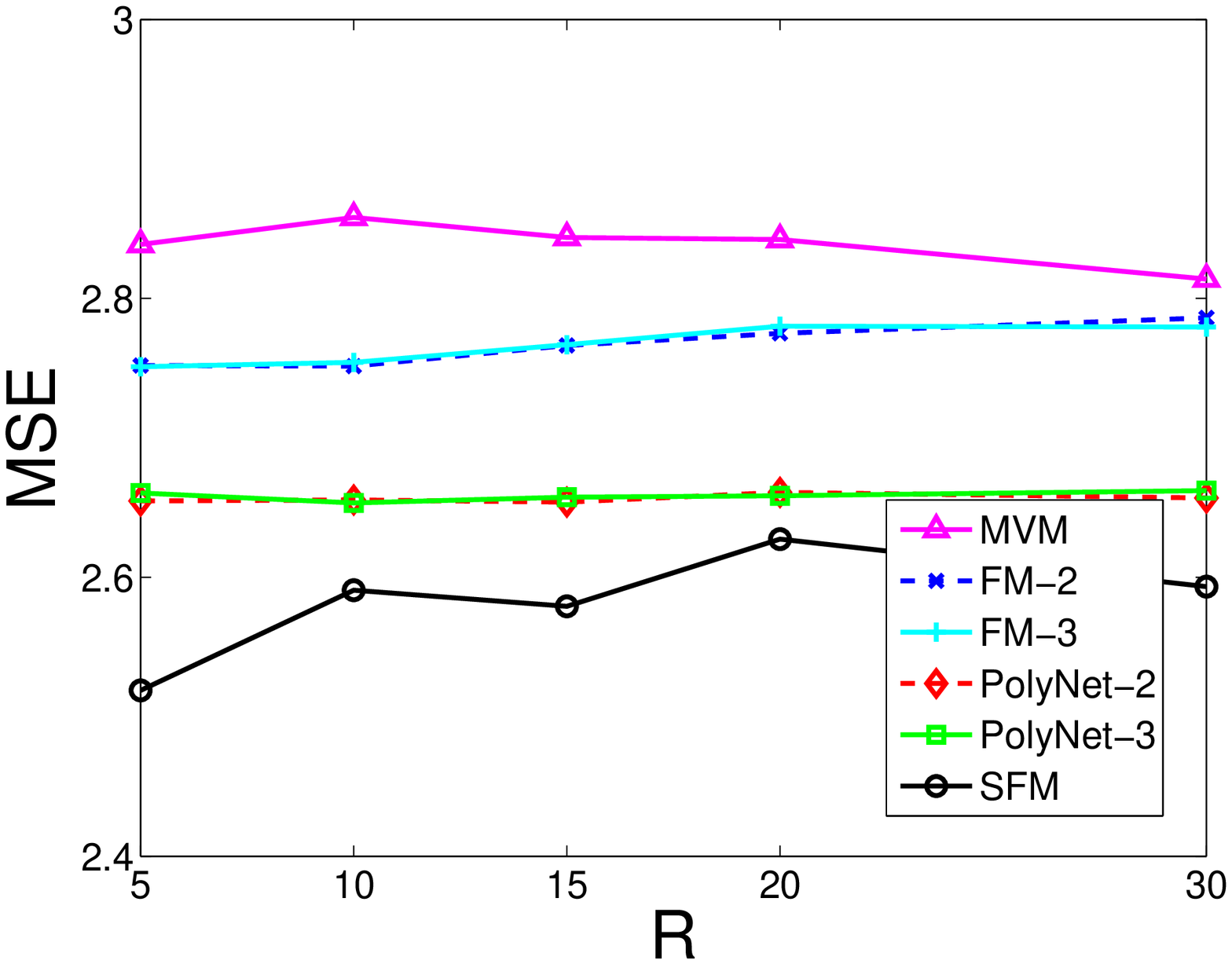}
    \end{minipage}
 }~
\caption{Sensitivity analysis of the latent dimension $R$.}\label{fig:exp_R}
\end{figure*}

\subsection{Performance Analysis}
The experimental results are shown in Table~\ref{tab:exp_mse}. The best method of each dataset is in bold.  For clarity, on the right of the tables we show the percentage improvement of the proposed SFM method over a variety of methods. From these results, we can observe that SFM consistently outperforms all the comparison methods. We also make a few comparisons and summarize our findings as follows. 

Compared with MF, SFM performs better with an average improvement of nearly 50\%. 
MF usually performs well in practice~\cite{ling2014ratings,rendle2012factorization}, while in datasets which are extremely sparse, as is shown in our case, MF is unable to learn an accurate representation of users/items. 
Thus MF under-performs other methods which takes the meta information into consideration. 

In both FM and PolyNet methods, the feature vectors from all the modes are concatenated as a single input feature vector. 
The major difference between these two methods is the choice of kernel applied~\cite{blondel2016higher}. The polynomial kernel used in PolyNet considers all monomials (the products of features), i.e., all combinations of features \textit{with} replacement. The ANOVA kernel used in FM considers only monomials composed of distinct features, i.e., feature combinations \textit{without} replacement. 
Compared with the best results obtained from FM methods and from PolyNet methods, SFM leads to an average improvement of 3.3\% and 2.4\% in MSE, respectively. 

The primary reason behind the results is how the latent factors of each feature are learned. 
For any factorization based method, the latent factors of a feature are essentially learned from its interactions with other features observed in the data, as can be observed from its update rule.  
In FM and PolyNet, all the feature interactions are taken into consideration without distinguishing the features from different modes. 
As a result, important feature interactions (e.g., the interactions between the given user and her friend) would be easily buried in irrelevant feature interactions from the same modes (e.g., the interactions between the friends of the same user). 
Hence, the learned latent factors are less representative in FM and PolyNet, compared with the proposed SFM. 
Besides, we can find that including higher-order interactions in FM and PolyNet (i.e., FM-3 and PolyNet-3) does not always improve the performance. 
Instead, it may even degrade the performance, as shown in Cloth, Yelp, and BX datasets. 
This is probably due to overfitting, as they need to include more parameters to model the interactions in higher orders while the datasets are extremely sparse such that the parameters cannot be properly learned.  

Compared to the MVM method, which models the full-order interactions among all the modes, our proposed SFM leads to an average improvement of 5.87\%. This is because not all the modes are relevant, and some irrelevant feature interactions may introduce unexpected noise to the learning task. The irrelevant information can even be exaggerated after combinations, thereby degrading performance. This suggests that preserving the nature of relational structure is important in building predictive models. 

\subsection{Computational Cost Analysis}

Next, we investigate the computational cost for comparison methods. The averaged training time (seconds per epoch) required for each dataset is shown in Fig.~\ref{fig:time}. 
We can easily find that the proposed SFM requires much less computational cost on all the datasets, 
especially for the Yelp dataset (roughly 11\% of computational cost required for training FM-3). 
The efficiency comes from the use of relational structure representation. As shown in Table~\ref{tab:dataset}, the number of non-zeros of the feature matrix $N_z(\mathbf{X})$ is much larger than the number of non-zeros of the relational structure representation $N_z(\mathcal{B})$. 
The amount of repeating patterns is much higher for the Yelp dataset than for the other dataset, because adding all the friends of a user significantly increases results in large repeating blocks in the plain feature matrix. 
Standard ML algorithms like the compared methods have typically at best a linear complexity in $N_z(\mathbf{X})$, while using the relational structure representation for SFM have a linear complexity in $N_z(\mathcal{B})$. 
This experiment substantiates the efficiency of the proposed SFM for large datasets.

\subsection{Analysis of the Impact of Data Sparsity}

We proceed by further studying the impact of data sparsity on different methods. 
As can be found in the experimental results, the improvement of SFM over the traditional collaborative filtering methods (e.g., MF) is significant for datasets that are sparse, 
mainly because the number of samples is too scarce to model the items and users adequately. 
We verify this finding by comparing the performance of comparison methods with MF on users with limited training data. 
Shown in Fig.~\ref{fig:exp_cold} is the gain of each method compared with MF for users with limited training samples, where  $G_1$, $G_2$, and $G_3$ are groups of users with $[1,3]$, $[4,6]$, and $[7,10]$ observed samples in the training set. 
Due to space limit, we only report the results from two Amazon datasets (Sport and Health) while the observations still hold for the rest datasets.  
It can be seen that the proposed SFM gains the most in group $G_1$, in which the users have extremely few training items. 
The performance gain starts to decrease with the number of training items available for each user. 
The results indicate that including meta information can be valuable information especially when limited information available. 

\subsection{Sensitivity analysis}
The number of latent factors $R$ is an important hyperparameter for the factorization models. 
We analyze different values of $R$ and report the averaged results in Fig.~\ref{fig:exp_R}. 
The results again show that SFM consistently outperforms other methods with various values of $R$. 
In contrast to findings in other related factorization models~\cite{yan2014coupled} where prediction error can steadily get reduced with larger $R$, we observe that the performance of each method is rather stable even with the increasing of $R$. 
It is reasonable in a general sense, as the expressiveness of the model is enough to describe the information embedded in data. Although larger $R$ renders the model with greater expressiveness, when the available observations regarding the target values are too sparse but the meta information is rich, only a few number of factors are required to fit the data well. 

%% file: table_dataset.tex
\begin{table*}[t]
 	\centering
	\caption{The statistics for each dataset. $N_z(X)$ and $N_z(\mathcal{B})$ are the number of non-zeros in plain formatted feature matrix and in relational structures, respectively. Game: Video Games, Cloth: Clothing, Shoes and Jewelry, Sport: Sports and Outdoors, Health: Health and Personal Care, Home: Home and Kitchen, Elec:  Electronics.}
	\label{tab:dataset}
	\begin{tabular}{lcccccccccc}
		\hline
		Dataset & \#Samples & \multicolumn{5}{c}{Mode} & Density & $N_z(X)$ & $N_z(\mathcal{B})$\\
		\hline
	    Amazon  & & \#Users & \#Items & \#Words & \#Categories & \#Links& \\
		\hline		
		Game & 231,780  & 24,303 & 10,672 & 7,500 & 193 & 17,974 & 0.089\% & 32.9M & 15.2M \\
		Cloth & 278,677 & 39,387 & 23,033 & 3,493 & 1,175 & 107,139 & 0.031\% & 25.6M & 7.3M\\		
		Sport & 296,337   & 35,598 & 18,357 & 5,202 & 1,432 & 73,040 & 0.045\% & 34.2M & 10.2M\\
		Health & 346,355  & 38,609 & 18,534 & 5,889 & 849 & 80,379 & 0.048\% & 33.6M & 12.1M\\
		Home   & 551,682   & 66,569 & 28,237 & 6,455 & 970 & 99,090 & 0.029\% & 46.8M & 19.4M\\
		Elec & 1,689,188  & 192,403 & 63,001 & 12,805 & 967 & 89,259 & 0.014\% & 161.5M & 69M\\
		\hline
		            &  & \#Users & \#Venues & \#Friends & \#Categories & \#Cities &  \\		
        \hline		            
		Yelp        & 1,319,870 & 88,009 & 40,520 & 88,009 & 892 & 412 & 0.037\% & 70.5M & 1.4M\\
		\hline		  
		            & & \#Users & \#Books & \#Countries & \#Ages & \#Authors  &\\		
        \hline		            
		BX          & 244,848 & 24,325 & 45,074 & 57 & 8 & 17,178 & 0.022\% & 1.2M & 163K\\
		\hline
	\end{tabular}
\end{table*}

%% file: table_regression.tex
\begin{table*}[t]
\centering
\caption{MSE comparison on all the datasets. The best results are listed in bold.}
\label{tab:exp_mse}
\begin{adjustbox}{max width=1\linewidth}
\begin{tabular}{lccccccc|ccc}
\hline
\multirow{2}{*}{Dataset}       & (a)                 & (b)                 & (c)                 & (d)                 & (e)                 & (f)                 & (g)                          & \multicolumn{3}{c}{Improvement of SFM verus} \\
                               & \textbf{MF}		& \textbf{MVM} & \textbf{FM-2}              & \textbf{FM-3}              & \textbf{PolyNet-2}           & \textbf{PolyNet-3}               & \textbf{SFM}                        & b  & min(c,d)       & min(e,f)       \\
                               \hline
Game & 1.569 $\pm$ 0.005  &	0.753 $\pm$ 0.007 &	0.764 $\pm$ 0.006 &	0.749 $\pm$ 0.007 &	0.749 $\pm$ 0.004 &	0.748 $\pm$ 0.006  &	\textbf{0.723 $\pm$ 0.006} &	4.06\%  & 3.52\% &	3.35\%  \\
Cloth & 1.624 $\pm$ 0.009  &	0.725 $\pm$ 0.046   &	0.678 $\pm$ 0.004 &	0.679 $\pm$ 0.004 &	0.678 $\pm$ 0.007 &	0.680 $\pm$ 0.005  &	\textbf{0.659 $\pm$ 0.013} & 9.03\%   &	2.82\% &	2.84\% \\
Sport & 1.290 $\pm$ 0.004  & 0.646 $\pm$ 0.019  &	0.638 $\pm$ 0.003 &	0.632 $\pm$ 0.007 &	0.631 $\pm$ 0.005 &	0.632 $\pm$ 0.005 &	\textbf{0.614 $\pm$ 0.011}  &	5.00\%  & 2.91\% &	2.79\% \\

Health & 1.568 $\pm$ 0.007 & 0.807 $\pm$ 0.012  & 0.779 $\pm$ 0.004 &	0.778 $\pm$ 0.004 &	0.779 $\pm$ 0.005 &	0.776 $\pm$ 0.005  &	\textbf{0.763 $\pm$ 0.019}  &	5.47\%  & 2.02\% &	1.77\% \\
Home & 1.591 $\pm$ 0.004   &	0.729 $\pm$ 0.067   &	0.714 $\pm$ 0.002 &	0.714 $\pm$ 0.004 &	0.690 $\pm$ 0.003 &	0.692 $\pm$ 0.005 &	\textbf{0.678 $\pm$ 0.008}  &	6.93\%  & 5.00\% &	1.72\% \\
Elec & 1.756 $\pm$ 0.002  &	0.792 $\pm$ 0.042    &	0.776 $\pm$ 0.006 &	0.749 $\pm$ 0.007 &	0.760 $\pm$ 0.004   &	0.757 $\pm$ 0.001  &	\textbf{0.747 $\pm$ 0.006}  &	5.69\%  & 0.27\% &	1.33\% \\
Yelp & 1.713 $\pm$ 0.003 & 1.2575 $\pm$ 0.013   & 1.277 $\pm$ 0.002 & 1.277 $\pm$ 0.002 & 1.272 $\pm$ 0.002	 & 1.272 $\pm$ 0.002  & \textbf{1.256 $\pm$ 0.010} & 0.09\%    & 1.58\%	& 1.19\% \\ 
BX & 4.094 $\pm$ 0.025  &	2.844 $\pm$ 0.024   &	2.766 $\pm$ 0.012 &	2.767 $\pm$ 0.014 &	2.654 $\pm$ 0.013 &	2.658 $\pm$ 0.013 &	\textbf{2.541 $\pm$ 0.025} &	10.66\%  & 8.16\% &	4.27\% \\
\hline
\multicolumn{3}{l}{Average on all datasets} & & & & &  & 5.87\% & 	3.29\% &	2.41\%\\
                               \hline
                               \hline
\end{tabular}
\end{adjustbox}
\end{table*}

%% file: 07-conclusion.tex
\section{Conclusions}\label{sec:conclusion}
In this paper, we introduce a generic framework for learning structural data from heterogeneous domains, which can explore the high order correlations underlying multi-view multi-way data. We develop structural factorization machines (SFMs) that learn the common latent spaces shared in the multi-view tensors while automatically adjust the contribution of each view in the predictive model. With the help of relational structure representation, we further provide an efficient approach to avoid unnecessary computation costs on repeating patterns of the multi-view data. 
It was shown that the proposed SFMs outperform state-of-the-art factorization models on eight large-scale datasets in terms of prediction accuracy and computational cost.

%% file: arxiv-main.bbl
\begin{thebibliography}{10}

\bibitem{acar2011all}
{\sc Acar, E., Kolda, T.~G., and Dunlavy, D.~M.}
\newblock All-at-once optimization for coupled matrix and tensor
  factorizations.
\newblock {\em arXiv preprint arXiv:1105.3422\/} (2011).

\bibitem{blondel2016higher}
{\sc Blondel, M., Fujino, A., Ueda, N., and Ishihata, M.}
\newblock Higher-order factorization machines.
\newblock In {\em Advances in Neural Information Processing Systems\/} (2016),
  pp.~3351--3359.

\bibitem{blondelIFU16}
{\sc Blondel, M., Ishihata, M., Fujino, A., and Ueda, N.}
\newblock Polynomial networks and factorization machines: New insights and
  efficient training algorithms.
\newblock In {\em Proceedings of the 33nd International Conference on Machine
  Learning\/} (2016), pp.~850--858.

\bibitem{cao2014tensor}
{\sc Cao, B., He, L., Kong, X., Yu, P.~S., Hao, Z., and Ragin, A.~B.}
\newblock Tensor-based multi-view feature selection with applications to brain
  diseases.
\newblock In {\em IEEE International Conference on Data Mining\/} (2014),
  pp.~40--49.

\bibitem{cao2017deepmood}
{\sc Cao, B., Zheng, L., Zhang, C., Yu, P.~S., Piscitello, A., Zulueta, J.,
  Ajilore, O., Ryan, K., and Leow, A.~D.}
\newblock Deepmood: Modeling mobile phone typing dynamics for mood detection.
\newblock In {\em Proceedings of ACM SIGKDD international conference on
  Knowledge discovery and data mining\/} (2017), pp.~747--755.

\bibitem{cao2016multi}
{\sc Cao, B., Zhou, H., Li, G., and Yu, P.~S.}
\newblock Multi-view machines.
\newblock In {\em ACM International Conference on Web Search and Data Mining\/}
  (2016), pp.~427--436.

\bibitem{cheng2016wide}
{\sc Cheng, H.-T., Koc, L., Harmsen, J., Shaked, T., Chandra, T., Aradhye, H.,
  Anderson, G., Corrado, G., Chai, W., Ispir, M., et~al.}
\newblock Wide \& deep learning for recommender systems.
\newblock In {\em DLRS\/} (2016), ACM, pp.~7--10.

\bibitem{chin2016libmf}
{\sc Chin, W.-S., Yuan, B.-W., Yang, M.-Y., Zhuang, Y., Juan, Y.-C., and Lin,
  C.-J.}
\newblock Libmf: A library for parallel matrix factorization in shared-memory
  systems.
\newblock {\em The Journal of Machine Learning Research 17}, 1 (2016),
  2971--2975.

\bibitem{covington2016deep}
{\sc Covington, P., Adams, J., and Sargin, E.}
\newblock Deep neural networks for youtube recommendations.
\newblock In {\em ACM Recommender Systems Conference (RecSys)\/} (2016), ACM,
  pp.~191--198.

\bibitem{duchi2011adaptive}
{\sc Duchi, J., Hazan, E., and Singer, Y.}
\newblock Adaptive subgradient methods for online learning and stochastic
  optimization.
\newblock {\em The Journal of Machine Learning Research 12\/} (2011),
  2121--2159.

\bibitem{guo2017deepfm}
{\sc Guo, H., Tang, R., Ye, Y., Li, Z., and He, X.}
\newblock Deepfm: A factorization-machine based neural network for ctr
  prediction.
\newblock {\em arXiv preprint arXiv:1703.04247\/} (2017).

\bibitem{guo2006mining}
{\sc Guo, H., and Viktor, H.~L.}
\newblock Mining relational data through correlation-based multiple view
  validation.
\newblock In {\em Proceedings of ACM SIGKDD international conference on
  Knowledge discovery and data mining\/} (2006), pp.~567--573.

\bibitem{harper2016movielens}
{\sc Harper, F.~M., and Konstan, J.~A.}
\newblock The movielens datasets: History and context.
\newblock {\em ACM Transactions on Interactive Intelligent Systems (TiiS) 5}, 4
  (2016), 19.

\bibitem{he2015delving}
{\sc He, K., Zhang, X., Ren, S., and Sun, J.}
\newblock Delving deep into rectifiers: Surpassing human-level performance on
  imagenet classification.
\newblock In {\em Proceedings of the IEEE international conference on computer
  vision\/} (2015), pp.~1026--1034.

\bibitem{he2014dusk}
{\sc He, L., Kong, X., Philip, S.~Y., Ragin, A.~B., Hao, Z., and Yang, X.}
\newblock Dusk: A dual structure-preserving kernel for supervised tensor
  learning with applications to neuroimages.
\newblock {\em matrix 3}, 1 (2014), 2.

\bibitem{he2016joint}
{\sc He, L., Lu, C.-T., Ma, J., Cao, J., Shen, L., and Yu, P.~S.}
\newblock Joint community and structural hole spanner detection via harmonic
  modularity.
\newblock In {\em Proceedings of the 22nd ACM SIGKDD International Conference
  on Knowledge Discovery and Data Mining\/} (2016), pp.~875--884.

\bibitem{he2017neural}
{\sc He, X., and Chua, T.-S.}
\newblock Neural factorization machines for sparse predictive analytics.
\newblock In {\em Proceedings of International ACM SIGIR Conference on Research
  and Development in Information Retrieval\/} (2017).

\bibitem{hinton2012rmsprop}
{\sc Hinton, G., Srivastava, N., and Swersky, K.}
\newblock Rmsprop: Divide the gradient by a running average of its recent
  magnitude.
\newblock {\em Neural networks for machine learning, Coursera lecture 6e\/}
  (2012).

\bibitem{huang2013learning}
{\sc Huang, P.-S., He, X., Gao, J., Deng, L., Acero, A., and Heck, L.}
\newblock Learning deep structured semantic models for web search using
  clickthrough data.
\newblock In {\em ACM International Conference on Information and Knowledge
  Management\/} (2013), ACM, pp.~2333--2338.

\bibitem{juan2016field}
{\sc Juan, Y., Zhuang, Y., Chin, W.-S., and Lin, C.-J.}
\newblock Field-aware factorization machines for ctr prediction.
\newblock In {\em Proceedings of the 10th ACM Conference on Recommender
  Systems\/} (2016), ACM, pp.~43--50.

\bibitem{kingma2014adam}
{\sc Kingma, D., and Ba, J.}
\newblock Adam: A method for stochastic optimization.
\newblock {\em arXiv preprint arXiv:1412.6980\/} (2014).

\bibitem{kolda2009tensor}
{\sc Kolda, T.~G., and Bader, B.~W.}
\newblock Tensor decompositions and applications.
\newblock {\em SIAM review 51}, 3 (2009), 455--500.

\bibitem{koren2008factorization}
{\sc Koren, Y.}
\newblock Factorization meets the neighborhood: a multifaceted collaborative
  filtering model.
\newblock In {\em Proceedings of ACM SIGKDD international conference on
  Knowledge discovery and data mining\/} (2008), pp.~426--434.

\bibitem{koren2010factor}
{\sc Koren, Y.}
\newblock Factor in the neighbors: Scalable and accurate collaborative
  filtering.
\newblock {\em ACM Transactions on Knowledge Discovery from Data (TKDD) 4}, 1
  (2010), 1.

\bibitem{liang2017icdm}
{\sc Liang, T., He, L., Lu, C.-T., Chen, L., Yu, P.~S., and Wu, J.}
\newblock A broad learning approach for context-aware mobile application
  recommendation.
\newblock In {\em 2017 IEEE International Conference on Data Mining (ICDM)\/}
  (Nov. 2017), pp.~955--960.

\bibitem{ling2014ratings}
{\sc Ling, G., Lyu, M.~R., and King, I.}
\newblock Ratings meet reviews, a combined approach to recommend.
\newblock In {\em Proceedings of the 8th ACM Conference on Recommender
  systems\/} (2014), pp.~105--112.

\bibitem{livni2014computational}
{\sc Livni, R., Shalev-Shwartz, S., and Shamir, O.}
\newblock On the computational efficiency of training neural networks.
\newblock In {\em Advances in Neural Information Processing Systems\/} (2014),
  pp.~855--863.

\bibitem{lu2017mfm}
{\sc Lu, C.-T., He, L., Shao, W., Cao, B., and Yu, P.~S.}
\newblock Multilinear factorization machines for multi-task multi-view
  learning.
\newblock In {\em Proceedings of the Tenth ACM International Conference on Web
  Search and Data Mining\/} (2017), pp.~701--709.

\bibitem{lu2016item}
{\sc Lu, C.-T., Xie, S., Shao, W., He, L., and Yu, P.~S.}
\newblock Item recommendation for emerging online businesses.
\newblock In {\em Proceedings of International Joint Conference Artificial
  Intelligence\/} (2016), pp.~3797--3803.

\bibitem{mcauley2013hidden}
{\sc McAuley, J., and Leskovec, J.}
\newblock Hidden factors and hidden topics: understanding rating dimensions
  with review text.
\newblock In {\em Proceedings of the 7th ACM conference on Recommender
  systems\/} (2013), pp.~165--172.

\bibitem{mcauley2015inferring}
{\sc McAuley, J., Pandey, R., and Leskovec, J.}
\newblock Inferring networks of substitutable and complementary products.
\newblock In {\em Proceedings of ACM SIGKDD international conference on
  Knowledge discovery and data mining\/} (2015), pp.~785--794.

\bibitem{novikov2016exponential}
{\sc Novikov, A., Trofimov, M., and Oseledets, I.}
\newblock Exponential machines.
\newblock In {\em International Conference on Learning Representations\/}
  (2017).

\bibitem{qu2016product}
{\sc Qu, Y., Cai, H., Ren, K., Zhang, W., Yu, Y., Wen, Y., and Wang, J.}
\newblock Product-based neural networks for user response prediction.
\newblock In {\em Data Mining (ICDM), 2016 IEEE 16th International Conference
  on\/} (2016), IEEE, pp.~1149--1154.

\bibitem{rendle2012factorization}
{\sc Rendle, S.}
\newblock Factorization machines with {libFM}.
\newblock {\em Intelligent Systems and Technology 3}, 3 (2012), 57.

\bibitem{rendle2013scaling}
{\sc Rendle, S.}
\newblock Scaling factorization machines to relational data.
\newblock In {\em Proceedings of the VLDB Endowment\/} (2013), vol.~6, VLDB
  Endowment, pp.~337--348.

\bibitem{rendle2010pairwise}
{\sc Rendle, S., and Schmidt-Thieme, L.}
\newblock Pairwise interaction tensor factorization for personalized tag
  recommendation.
\newblock In {\em Proceedings of the third ACM international conference on Web
  search and data mining\/} (2010), ACM, pp.~81--90.

\bibitem{shan2016deep}
{\sc Shan, Y., Hoens, T.~R., Jiao, J., Wang, H., Yu, D., and Mao, J.}
\newblock Deep crossing: Web-scale modeling without manually crafted
  combinatorial features.
\newblock In {\em Proceedings of ACM SIGKDD international conference on
  Knowledge discovery and data mining\/} (2016), ACM, pp.~255--262.

\bibitem{singh2008relational}
{\sc Singh, A.~P., and Gordon, G.~J.}
\newblock Relational learning via collective matrix factorization.
\newblock In {\em Proceedings of ACM SIGKDD international conference on
  Knowledge discovery and data mining\/} (2008), pp.~650--658.

\bibitem{wang2017deep}
{\sc Wang, R., Fu, B., Fu, G., and Wang, M.}
\newblock Deep \& cross network for ad click predictions.
\newblock {\em arXiv preprint arXiv:1708.05123\/} (2017).

\bibitem{xiao2017attentional}
{\sc Xiao, J., Ye, H., He, X., Zhang, H., Wu, F., and Chua, T.-S.}
\newblock Attentional factorization machines: Learning the weight of feature
  interactions via attention networks.
\newblock In {\em International Joint Conference on Artificial Intelligence\/}
  (2017).

\bibitem{xu2013survey}
{\sc Xu, C., Tao, D., and Xu, C.}
\newblock A survey on multi-view learning.
\newblock {\em arXiv:1304.5634\/} (2013).

\bibitem{yan2014coupled}
{\sc Yan, L., Li, W.-j., Xue, G.-R., and Han, D.}
\newblock Coupled group lasso for web-scale {CTR} prediction in display
  advertising.
\newblock In {\em International Conference on Machine Learning\/} (2014),
  pp.~802--810.

\bibitem{zhang2017connecting}
{\sc Zhang, J., Lu, C.-T., Cao, B., Chang, Y., and Yu, P.~S.}
\newblock Connecting emerging relationships from news via tensor factorization.
\newblock In {\em Proceedings of IEEE International Conference on Big Data\/}
  (2017), IEEE.

\bibitem{zhang2016deep}
{\sc Zhang, W., Du, T., and Wang, J.}
\newblock Deep learning over multi-field categorical data.
\newblock In {\em European conference on information retrieval\/} (2016),
  Springer, pp.~45--57.

\bibitem{zheng2017joint}
{\sc Zheng, L., Noroozi, V., and Yu, P.~S.}
\newblock Joint deep modeling of users and items using reviews for
  recommendation.
\newblock In {\em Proceedings of the Tenth ACM International Conference on Web
  Search and Data Mining\/} (2017), pp.~425--434.

\bibitem{zhou2017deep}
{\sc Zhou, G., Song, C., Zhu, X., Ma, X., Yan, Y., Dai, X., Zhu, H., Jin, J.,
  Li, H., and Gai, K.}
\newblock Deep interest network for click-through rate prediction.
\newblock {\em arXiv preprint arXiv:1706.06978\/} (2017).

\bibitem{zhu2017deep}
{\sc Zhu, J., Shan, Y., Mao, J., Yu, D., Rahmanian, H., and Zhang, Y.}
\newblock Deep embedding forest: Forest-based serving with deep embedding
  features.
\newblock In {\em Proceedings of ACM SIGKDD international conference on
  Knowledge discovery and data mining\/} (2017).

\end{thebibliography}
